# Visual Language Models show widespread visual deficits on neuropsychological tests

Gene Tangtartharakul, Katherine R. Storrs

School of Psychology, The University of Auckland

**Abstract**

Visual Language Models (VLMs) show remarkable performance in visual reasoning tasks, successfully tackling college-level challenges that require high-level understanding of images. However, some recent reports of VLMs struggling to reason about elemental visual concepts like orientation, position, continuity, and occlusion suggest a potential gulf between human and VLM vision. Here we use the toolkit of neuropsychology to systematically assess the capabilities of three state-of-the-art VLMs across visual domains. Using 51 tests drawn from six clinical and experimental batteries, we characterise the visual abilities of leading VLMs relative to normative performance in healthy adults. While the models excel in straightforward object recognition tasks, we find widespread deficits in low- and mid-level visual abilities that would be considered clinically significant in humans. These selective deficits, profiled through validated test batteries, suggest that an artificial system can achieve complex object recognition without developing foundational visual concepts that in humans require no explicit training.

## Introduction

Counterintuitively, the mental abilities that seem simplest to humans are often the hardest to achieve in artificial intelligence (AI) — a fact known as Moravec's Paradox (Moravec, 1988). The most well-known example is the relative ease of implementing abstract reasoning in computers, compared to perception or sensorimotor control. Here we document a paradox within perception itself: state-of-the-art AI systems appear to have mastered "complex" visual tasks before they are competent at "simple" ones.

Deep Neural Networks (DNNs) are biologically inspired artificial systems with hierarchical architectures loosely resembling sensory cortices. Convolutional DNNs, in particular, have been extensively evaluated as models of mammalian visual cortex, since both systems share spatially-restricted "receptive fields" at early stages of processing. Convolutional DNNs are widely considered the best available models of neural responses in the brain's ventral stream (Doerig et al., 2023; Kriegeskorte, 2015; Marblestone et al., 2016; Storrs & Kriegeskorte, 2019; Yamins & DiCarlo, 2016), and can be trained to perform diverse perceptual tasks. However, their limited generalisation ability means they usually require fine-tuning on new tasks (Biscione et al., 2024; Puebla & Bowers, 2022), and they struggle with adversarial robustness, occlusion and 3D understanding (Abbas & Deny, 2022; Alcorn et al., 2019; Bowers et al., 2023).

Meanwhile, Large Language Models (LLMs) have come to prominence in the past five years. Using transformer (Vaswani et al., 2023) architectures and training on vast text corpora, LLMs can generate human-like text responses and conversations. Their natural-language interface makes it possible to administer cognitive tests originally designed to assess thinking, problem-solving, and decision-making skills in humans, providing a wealth of tools to evaluate artificial reasoning (Binz et al., 2024; Cherian et al., 2024; Ichien et al., 2024). The recent advent of Visual Language Models (VLMs) — multimodal transformer architectures capable of processing both text and images — now heralds a new era of general-purpose artificial vision systems.

VLMs encode multi-modal data into a shared representational space, merging linguistic and visual reasoning (Alayrac et al., 2022; Lu et al., 2019; Radford et al., 2021). Their ability to perform visual question answering (VQA) allows human-like interactions — we can ask questions of them as if they were a participant in a psychology clinic or experiment (Binz & Schulz, 2023). VLMs exhibit strong zero-shot learning, recognizing novel images without task-specific training (Radford et al., 2021; Wu et al., 2024) — a marked challenge for previous DNNs. They excel in diverse recognition tasks from face and landmark identification, to medical image interpretation and college-level reasoning over subject-specific diagrams, charts, and tables (Yue et al., 2024).

Despite their strengths, reports of idiosyncratic weaknesses in VLMs' visual abilities have also emerged. A recent assessment of VLMs using visual reasoning tasks from cognitive science found that VLMs underperform humans in the domains of intuitive physics, causal reasoning and intuitive psychology (Buschoff et al., 2025). Reports from computer science have revealed counterintuitive failures when VLMs are asked to reason about simple visual displays. For example, VLMs appear to struggle with visual arithmetic tasks such as counting line intersections, grids, or nested squares (Rahmanzadehgervi et al., 2024), as well as comparing object lengths or orientations (Huang et al., 2025; Z. Wang et al., 2024). Occlusions impact VLMs' understanding of objects and text in images (Qiu & Di, 2024; Rahmanzadehgervi et al., 2024; Yang & Di, 2024), and VLMs' interpretations of spatial context, such as relative position, orientation, direction, and perspective can be unreliable (Tong et al., 2024). This poor spatial reasoning about simple displays is surprising, given VLMs' excellence at much more complex-seeming tasks, and is particularly puzzling given their proficiency at interpreting images of documents, charts, and tables (Deng et al., 2024; Masry et al., 2022; Mathew et al., 2021).

While such findings suggest fundamental differences between human and VLM vision, no attempt has yet been made to systematically assess VLMs' visual function relative to human perception. Such an assessment would require direct comparisons of human and VLM performance on the same visual tests. Fortunately, a wealth of suitable tests and normative data already exists. Neuropsychological assessments are the traditional toolkit used by clinicians to assess an individual's perceptual and cognitive abilities. Any shortcomings in the performance on each test set, relative to a normative sample of healthy adults, can be used to build an individual's unique deficit profile. Outside of the clinic, experimental psychologists use conventional psychophysical measures with carefully controlled stimuli to isolate and measure particular aspects of visual processing (Bowers et al., 2023). Clinical and experimental psychology therefore provide an ideal set of standardised test batteries, often with normative performance data, to pinpoint and quantify disparities or similarities between machine and human perception across diverse visual processes. Neuropsychological assessments evaluate different stages of visual processing, with widely used tests like the Birmingham Object Recognition Battery (BORB) (Humphreys et al., 1993), offering tests of low, mid, and high-level vision, including semantic naming tasks. The Visual Object and Space Perception Battery (VOSP) (Warrington & James, 1991) meanwhile primarily focuses on visuospatial and object perception, while newer tests like the Leuven Perceptual Organization Test (L-POST) (Torfs et al., 2014) were especially designed to assess mid-level vision. There are also domain-specific tests, such as the Cambridge Face Memory Tests (CFMT) which assesses high-level face perception. Tasks from experimental psychology offer a complementary toolkit, having been designed to test specific hypotheses, and having well-established “typical” patterns of human responses (Biscione et al., 2024).

With these tools, the current study employed both clinical vision assessments and visual tests from experimental psychology to compare human and VLM visual function. We evaluated three state-of-the-art commercial VLMs with a comprehensive battery of tests spanning the visual processing hierarchy. Human normative data were either retrieved from neuropsychological tests or collected in online experiments. We aimed to explore whether the shortcomings of VLMs can be understood in terms of visual deficits in specific domains and to clearly delineate the VLM-human perceptual gap.

## Results

We evaluated three VLMs on 51 visual perception tests, sourced from six batteries (**Figure 1a**). We chose VLMs that were state-of-the-art at the time of testing (August-November 2024): *GPT4o (2024-05-13)*, *Claude-3.5 Sonnet (new)*, and *Gemini-1.5 pro*, and which have achieved outstanding performance across multiple visual reasoning benchmarks (Duan et al., 2024). Of the 51 perceptual tests, 29 were sourced from one of six validated neuropsychological batteries used to assess vision in clinical examinations and research: the *Birmingham Object Recognition Battery* (BORB) (Humphreys & Riddoch, 1993); the *Hooper Visual Organization Test* (VOT) (Hooper, 1983); the *Leuven Perceptual Organization Test* (L-POST) (Torfs et al., 2014); the *Leuven Embedded Figure Test* (L-EFT) (de-Wit et al., 2017); and the *Dartmouth Face Perception Test* (DFPT) (Dalrymple et al., 2017). We supplemented these with 22 tests from the *MindSet:Vision* toolbox (Biscione et al., 2024). *MindSet:Vision* is a battery of perceptual tests based on well-established results in experimental psychology, designed to comprehensively evaluate whether a model shows signatures of human-like perception. Each test consisted of 5-128 trials of unique image stimuli, so that each model was probed in total with 1586 stimuli. Importantly, the majority of the image stimuli used to probe the models could not have appeared in their training datasets, since the stimuli were either digitised by us from proprietary clinical batteries not available online (6 tests) or procedurally generated by us following test specifications (22 tests).

All tests used one of four task formats: a match-to-sample task, a same-different task, a yes/no task, or a naming task (**Figure 1a**). We provided the models with instructions and image stimuli via their API, recording outputs for analysis. We submitted one API request for each trial, so that trial behaviour was not influenced by the context of previous trials. Example VLM exchanges for each task type are given in **Figure 1b**, and full prompt text for each task type and test is provided in the Method and Supplementary Method, respectively. For clinical tests, we formatted instruction prompts as closely as possible to how their source batteries stated they should be administered in clinical settings. We adapted these prompts for the psychological tests, retaining similar phrasing and structure. We included additional statements instructing the model to provide a single definitive answer, even if it was uncertain. This approach mimics clinical testing scenarios, where patients may be encouraged to rely on guesses or intuition when unsure.

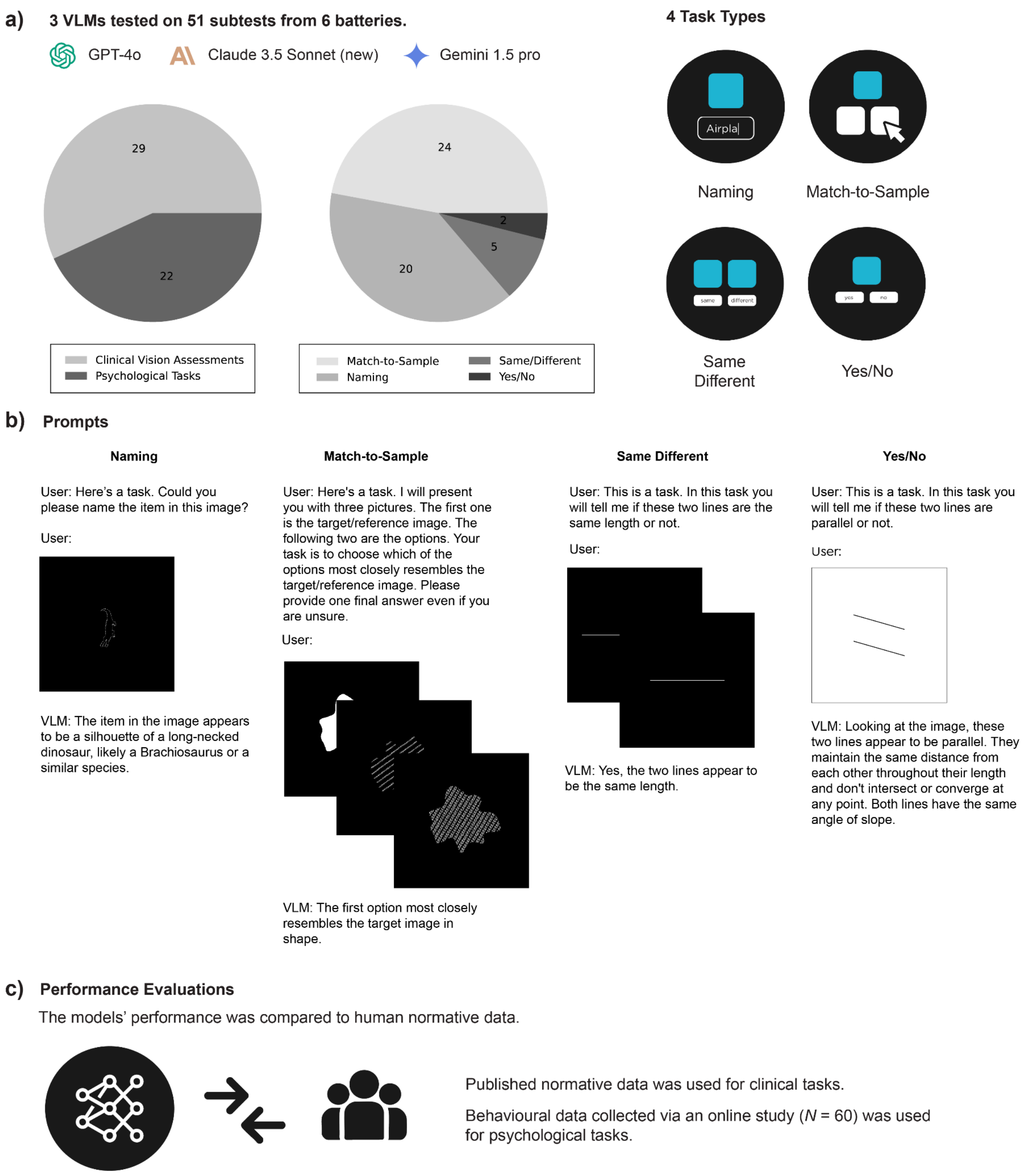

**Figure 1: Study workflow. a)** Three commercial VLMs were tested on 51 tests derived from six test batteries. These tests fall into four task categories: naming, match-to-sample, same-different, and yes/no tasks. **b)** Example trials for each task type using MindSet stimuli. The naming task prompts the model to identify the item in an image. In the same-different task, the model determines whether a specific property of two images is identical. The match-to-sample task asks the model to select the image most similar to a target. Depending on the test, two or three image options may be presented. The yes/no task evaluates whether a condition is true, for instance, two lines being parallel. **c)** Model accuracy was compared to human normative data. Published normative data was used for clinical tests, and a new behavioural dataset collected via an online study was used for psychological tests.

We compared VLM accuracy on a test-by-test basis to normative accuracy in healthy adults (**Figure 1c**). For 24 of the tests, published human normative accuracy data were available from the source battery. For the remaining 27 tests (predominantly psychological tests from the *MindSet:Vision* toolbox), we collected new human normative data via an online behavioural experiment (*N* = 60, ages 18-63 with no visual disturbances; see Method).

*VLMs can perform all tests, but show profound visual deficits compared to humans*

Performance was overall high, and substantially better than chance (mean accuracy across all models and tests = 69.51% +/- 24.79% SD, see Supplementary Figure S1). Below-chance accuracy was observed only in a handful of exceptions: GPT-4o performed below chance on five tests, Claude 3.5-Sonnet on three, and Gemini 1.5-pro on four. All models shared two below-chance tests: L-POST's hidden contour shape matching and MindSet's occluded shape matching. Since task types differed in their chance level, and individual tests differed in their complexity, we expressed model performance for each test relative to human normative accuracy in that test (see **Figure 2**).

When administered clinically, a patient is considered to show clinically significant impairment when their test performance falls more than two standard deviations (Jacobson & Truax, 1991; Kendall & Grove, 1988) below the normative healthy human mean. Across 50 tests (excluding HVOT, for which normative standard deviation was not available), GPT-4o and Claude 3.5-Sonnet showed clinically significant visual deficits on 18 tests, while Gemini 1.5-pro did so on 16 tests (see tests marked with asterisks in **Figure 2**). Conversely, we also identified super-human performance, defined as scores exceeding the normative mean by more than two standard deviations. GPT-4o and Claude's 3.5-Sonnet exceeded this criterion on one test, and Gemini 1.5-pro did so on two tests.

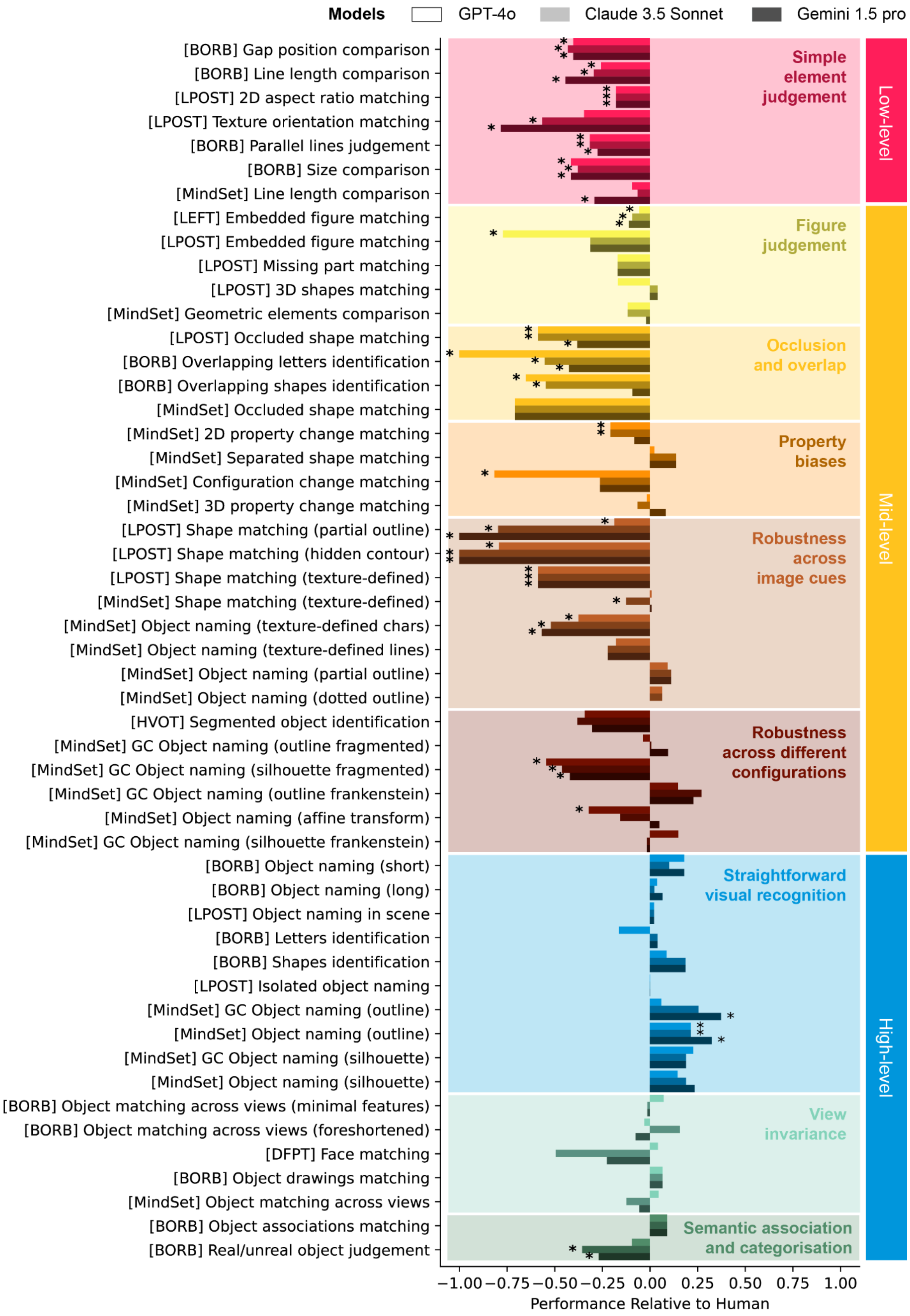

Models
GPT-4o
Claude 3.5 Sonnet
Gemini 1.5 pro
[BORB] Gap position comparison
[BORB] Line length comparison
[LPOST] 2D aspect ratio matching
[LPOST] Texture orientation matching
[BORB] Parallel lines judgement
[BORB] Size comparison
[MindSet] Line length comparison
[LEFT] Embedded figure matching
[LPOST] Embedded figure matching
[LPOST] Missing part matching
[LPOST] 3D shapes matching
[MindSet] Geometric elements comparison
[LPOST] Occluded shape matching
[BORB] Overlapping letters identification
[BORB] Overlapping shapes identification
[MindSet] Occluded shape matching
[MindSet] 2D property change matching
[MindSet] Separated shape matching
[MindSet] Configuration change matching
[MindSet] 3D property change matching
[LPOST] Shape matching (partial outline)
[LPOST] Shape matching (hidden contour)
[LPOST] Shape matching (texture-defined)
[MindSet] Shape matching (texture-defined)
[MindSet] Object naming (texture-defined chars)
[MindSet] Object naming (texture-defined lines)
[MindSet] Object naming (partial outline)
[MindSet] Object naming (dotted outline)
[HVOT] Segmented object identification
[MindSet] GC Object naming (outline fragmented)
[MindSet] GC Object naming (silhouette fragmented)
[MindSet] GC Object naming (outline frankenstein)
[MindSet] Object naming (affine transform)
[MindSet] GC Object naming (silhouette frankenstein)
[BORB] Object naming (short)
[BORB] Object naming (long)
[LPOST] Object naming in scene
[BORB] Letters identification
[BORB] Shapes identification
[LPOST] Isolated object naming
[MindSet] GC Object naming (outline)
[MindSet] Object naming (outline)
[MindSet] GC Object naming (silhouette)
[MindSet] Object naming (silhouette)
[BORB] Object matching across views (minimal features)
[BORB] Object matching across views (foreshortened)
[DFPT] Face matching
[BORB] Object drawings matching
[MindSet] Object matching across views
[BORB] Object associations matching
[BORB] Real/unreal object judgement
Simple element judgement
Figure judgement
Occlusion and overlap
Property biases
Robustness across image cues
Robustness across different configurations
Straightforward visual recognition
View invariance
Semantic association and categorisation
Low-level
Mid-level
High-level
−1.00 −0.75 −0.50 −0.25 0.00 0.25 0.50 0.75 1.00
Performance Relative to Human

**Figure 2: Model performance.** The models' performance relative to humans across all 51 tests. The models' accuracy was normalized relative to humans' normative accuracy to obtain the relative performance values. A relative performance of zero indicates that a model did as well as humans on a test. Negative and positive values indicate that the model performance was worse or better than humans, respectively. Tests are categorized into low-, mid-, or high-level visual assessments and further grouped into nine processes reflecting finger-grained visual abilities. Asterisks denote clinically significant deviations from human performance (>± 2 standard deviations from the normative mean).

*VLMs have similar strengths and weaknesses to one another*

GPT4o, Claude 3.5-Sonnet, and Gemini 1.5-pro exhibited highly correlated performance across 51 tests (all $\rho(49) > .79$, $p < .001$; **Figure 3a**), demonstrating a similar profile of strengths and weaknesses across these challenges. Permutation tests were performed to compare mean test scores between each model pair, followed by correction for multiple comparison. Comparing normalized scores between the three models suggested that none of the three models had a more human-like performance profile than any other ($p > .05$; **Figure 3b**). For this reason, we chose to average across the three VLMs, and focus on the average normalised performance across all models in subsequent analyses.

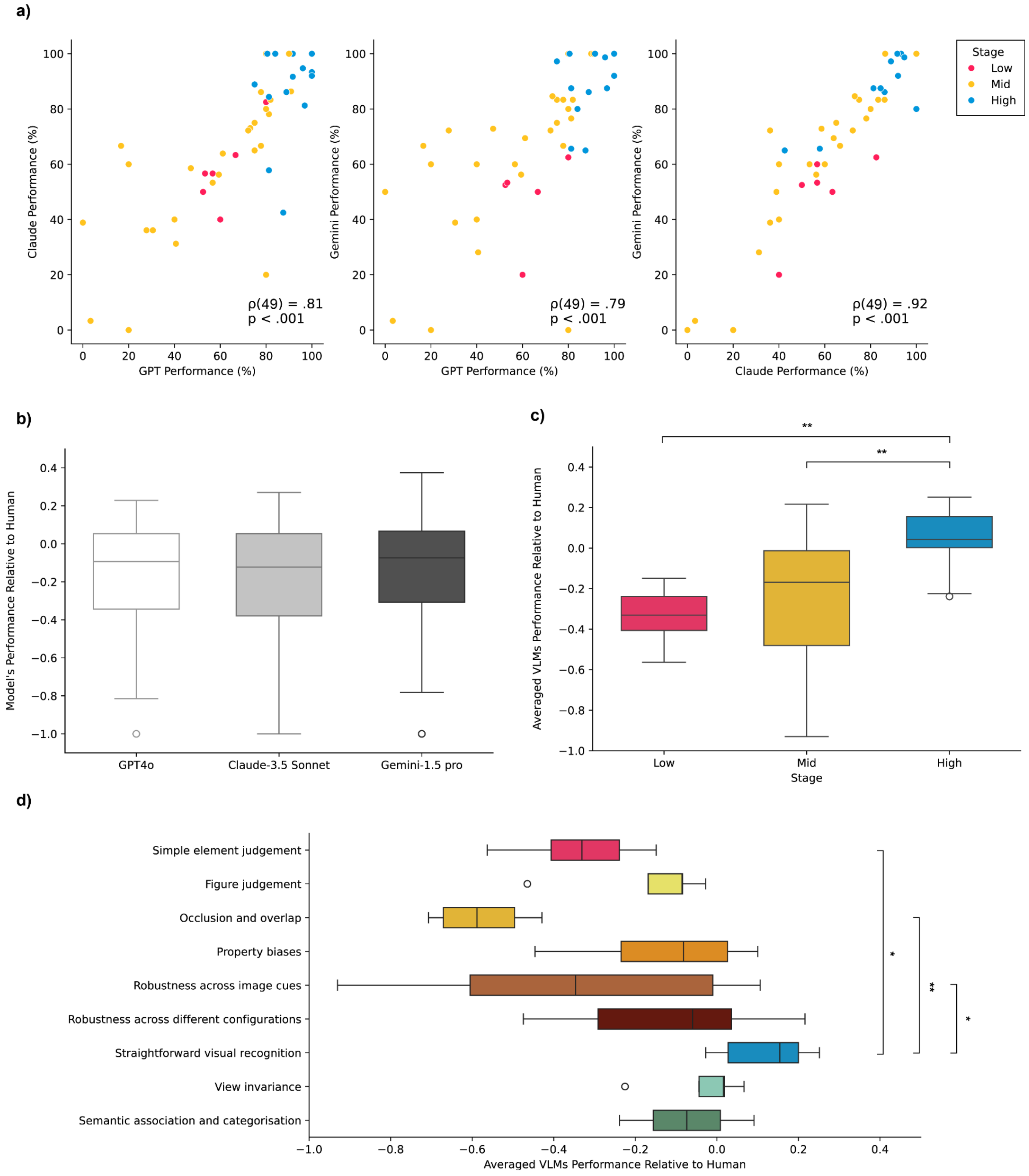

**Figure 3: Performance broken down by model, visual stage, and visual process. a)** Highly correlated performance among three models (left; GPT4o vs. Claude3.5-Sonnet, middle; GPT4o vs. Gemini-1.5 pro, right; Claude3.5-Sonnet vs. Gemini1.5-pro) **b)** Performance across tests, relative to human accuracy, for each model. **c)** Performance relative to human accuracy, grouping tests according to the visual processing stage they predominantly probe, aggregated across all VLMs. **d)** Relative performance, grouping tests into nine finer-grained categories according to which visual process they predominantly probe, aggregated across all VLMs. Pairwise comparisons were corrected using Bonferroni corrections. Asterisks denote statistical significance at the following Bonferroni-corrected levels: * .05, ** .01, *** < .001, based on a permutation test of group means.

*VLMs only excel at straightforward object-recognition, with deficits in low- and mid-level vision*

The 51 tests tapped a diverse range of visual processes and abilities, across all stages of the visual hierarchy. In order to characterise the strengths and weaknesses of the models, we grouped the tests at a fine grain (according to the main visual ability tapped by each test) and a coarse grain (according to which visual processing stage this ability belonged to). We made these designations using a combination of the test creator's intent, the opinion of the authors, and opinions informally canvassed from another 11 professional vision scientists. Our designations of process and stage are by no means intended as precise or ground-truth descriptions of the tests. They serve merely to group similar tests together and to provide some intuition about the sorts of visual intelligence required to do well on them.

At a coarse grain, we categorized each test into one of three visual processing stages: low-, mid-, or high-level. Low-level visual processing is associated with early retinotopic visual areas in humans (Bosking et al., 2002; Kamitani & Tong, 2005; Schwarzkopf & Rees, 2013; Sperandio & Chouinard, 2015; Victor et al., 1994). It involves the extraction and comparison of basic image components such as edges, shape, size, and orientation. We assessed low-level vision using selected tests from the Birmingham Object Recognition Batteries (BORB) (Humphreys & Riddoch, 1993) and the Leuven Perceptual Organization Test (L-POST) (Torfs et al., 2014) which involve evaluating simple image features, e.g. the length or orientation of lines. Mid-level processing is associated with perceptual organization (Anderson, 2020; Koffka, 1935), utilizing proximity, continuity, and other figure-ground segregation clues to form coherent representations (Persike & Meinhardt, 2016). We probed mid-level vision using L-POST (Torfs et al., 2014) and the Leuven Embedded Figure Test (L-EFT) (de-Wit et al., 2017), which were designed to assess perceptual organization without requiring semantic knowledge. We also included the Hooper Visual Organization Test (HVOT) (Hooper, 1983) and several tests from the *MindSet: Vision* (Biscione et al., 2024) battery that assess instances where mid-level processes are required for successful object identification (e.g. recognising a texture-defined object). High-level processing combines visual representations with semantic knowledge for object identification, categorization, and generalization (Cox, 2014; Kandel et al., 2014; Ullman, 2000). We used BORB and *MindSet: Vision* to assess the models' object identification across various object views and image styles.

On average, the VLMs slightly outperformed humans in high-level recognition tests but substantially underperformed humans in both low-level and mid-level visual tests (**Figure 3c**). Pairwise permutation tests comparing human-relative performance between different visual stages showed that VLMs performed worse in tests of low- (mean difference = 0.39, $p = .003$) and mid-level vision ($MD = 0.32$, $p < .001$) compared to high-level visual tests. The VLMs' relative performance did not differ between low- and mid-level visual tests ($MD = 0.07$, $p = .554$) (**Figure 3c**). Paradoxically, the tests that seem simplest to humans — like comparing the orientation of lines, or matching shapes drawn in different line styles — seem hardest for VLMs.

We further divided the three stages into nine finer-grained categories based on specific visual abilities assessed by the tests. The first category, *simple element judgement*, involves comparing primitive visual features such as size, length, and orientation. The second category, *figure judgement*, involves matching and comparing abstract shapes, such as blobs and embedded figures. The third category, *occlusion and overlap,* groups together tests that require resolving occlusion to answer correctly. Tests in the fourth category– *property biases*– assess whether a model exhibits biases toward certain image features in a manner similar to humans (e.g. a tendency to attend to changes in "non-accidental properties" rather than equal metric changes to a shape) (Amir et al., 2012, 2014). The fifth and sixth categories probe whether a model shows *robustness across image cues* (such as the texture or outline style used to define a figure), and *robustness across different configurations* (such as fragmented or rearranged parts) when recognising shapes or objects. The final three categories subdivide high-level object recognition tests into those that involve *straightforward visual recognition*, those requiring *view invariance*, and those requiring contextual or *semantic association and categorisation* of objects. For an illustrated list of all tests, along with their coarse and fine groupings, see **Table S1** in the **Supplementary Methods**.

Using these finer-grained distinctions, we similarly investigated whether the VLMs performed better or worse than humans in a specific test group (**Figure 3d**). The VLMs performed significantly better, relative to humans, on tests that involved straightforward object recognition (e.g., outlines or silhouettes of objects, shapes, and letters) than those involving the judgment of simple visual elements (e.g., line lengths and circle sizes) ($MD = 0.46$, $p = .001$), overlap or occlusion ($MD = 0.70$, $p < .001$), and those requiring robustness across different image cues (*MD* = 0.46, *p* < .001) (**Figure 3d**). No other comparisons were significant after adjusting for multiple comparisons.

## Discussion

We challenged three commercial VLMs on 51 visual perception tests taken from a combination of clinical (29) and psychological (22) batteries. Our test batteries extensively covered low-, mid-, and high-level visual stages, and tapped into diverse visual processes. This approach allowed us to systematically profile the strengths and weaknesses of current VLMs. All three VLMs shared the same clear profile: they matched or outperformed humans in straightforward high-level visual recognition tests but showed profound difficulties on most tests of low-, and mid-level visual abilities.

Completing low-level visual tasks like position, orientation, or size comparisons relies on basic feature extraction to detect physical changes in the stimuli. While VLMs have shown an understanding of 2D geometry, such as parallelism and orthogonality (Azad et al., 2024), our findings suggest otherwise. This discrepancy may arise from differences in assessment format (multiple-choice vs. close-ended questions) and the use of colour grounding in the prompt stimulus and instructions. Success in these tasks requires understanding relative concepts– referencing and comparing objects– an evident shortcoming of VLMs (Huang et al., 2025; Tong et al., 2024). Their difficulty may stem from these properties being implicit and rarely captured in image-text training data. However, a recent study on open-source models argued that, although pre-trained vision encoders capture sufficient representations of these relationships, text decoders struggle to leverage them effectively without further fine-tuning (Huang et al., 2025).

Mid-level tests require perceptual organization to recognize figures or objects. For example, L-POST matching tasks require grouping of contours or textures based on proximity, continuity, or closure, while MindSet naming tasks require resolving partial contours in order to name objects. While perceptual organization in VLMs remains underexplored, recent findings show state-of-the-art VLMs are unable to trace line paths, implying their failure to resolve continuity (Rahmanzadehgervi et al., 2024). The models' limited perceptual organization is also evident in their poor performance in cases involving overlaps and occlusions. Although the models matched human performance in naming normally-presented letters and shapes (*BORB letters identification* and *BORB shapes identification*), their accuracy declined considerably when the same stimuli overlapped (*BORB overlapping letters* and *shapes identification*). This agrees with prior findings on VLMs' difficulty identifying occluded objects (Qiu & Di, 2024; Yang

& Di, 2024). Simply having a circle superimposing a word has been shown to impair letter recognition, and overlapping shapes can similarly impair the models' counting accuracy (Rahmanzadehgervi et al., 2024).

The VLMs we tested performed similarly poorly in low- and mid-level visual tasks. While some mid-level tests involve high-level object knowledge, the specialized mid-level batteries L-POST and L-EFT use abstract figures to better isolate visual from semantic abilities. Abstract shapes are likely rare in the models' training data and, more importantly, difficult to describe semantically with text captions. These tasks could therefore be considered out-of-distribution, and challenge the models to zero-shot generalisation.

Despite impaired low- and mid-level performance, VLMs matched or even outperformed humans in high-level tasks. They easily identify ImageNet objects with simplified outlines or silhouettes in basic naming tasks, despite mixed evidence on their zero-shot robustness to distribution shifts (Radford et al., 2021; Wang et al., 2024). They also performed well at recognizing an object across different 3D views. A key factor contributing to the models' success in high-level visual tests is likely the presence of nameable shapes, letters, and objects– the domain in which VLMs excel (Yang et al., 2023) – as opposed to abstract shapes in low- and mid-level tests. Although some mid-level tasks also included nameable objects, they were subject to manipulations like rotation, offsets, and fragmentation (subgrouped as *robustness across image cues* and *robustness across different configurations*), which likely hindered VLMs' ability to name them. Unlike straightforwardly-rendered objects, those with mid-level modifications required resolving partial image information before identification.

While the super-human object recognition accuracy for isolated object images may be surprising, high performance in VLMs is attributable to, first, a broad knowledge base as they are trained on vast datasets of images and texts from the internet. This makes them excel at drawing fine-grained distinctions between similar objects, once trained. Second, even when humans have equivalent visual understanding and object knowledge, they may struggle to find the correct word, resulting in missing answers, semantic errors, or circumlocution, which hinder task performance. VLMs are also unhampered by factors like fatigue or drifting attention which may reduce human performance.

The profile of strengths and deficits documented here reveals which visual concepts can and cannot be learned well from web-based image and text pairings. Foundation models like commercial VLMs are exposed to vast amounts of data from web-based corpora (e.g., LAION, Common Crawl) to learn relationships between images and their corresponding textual descriptions by embedding them within a shared multimodal representational space (Bordes et al., 2024; Jung, 2023; Lu et al., 2019; Radford et al., 2021; Tsimpoukelli et al., 2021). For VLMs to perform well in low- and mid-level tasks, they would likely need to be exposed to training images paired with detailed textual descriptions of local features and their configurations. While this may seem obvious, it highlights a stark contrast between human and VLM vision. When a human encounters the neuropsychological and experimental tests here for the first time, they are likely to seem trivial, rather than being 'specialized' or 'domain-specific' tasks on which we need to specifically train. Through learning the overarching objective of navigating the visible world (Groen et al., 2017), our visual systems have learned a general ability to flexibly make *ad hoc* judgements about novel visual elements and configurations. In contrast, observed VLMs failures on fundamental tasks might reflect the models' inability to manage the binding problem (Campbell et al., 2024; Greff et al., 2020) — difficulty in simultaneously representing multiple objects and their associated features without interference (Treisman, 1996).

If the visual deficits documented here do in fact arise from limitations in the approach of learning image and text embeddings from available human-produced data, then they are unlikely to be fixed by incremental updates to VLMs trained using this approach. However, we do not wish to suggest that the shortcomings identified here are insurmountable, nor that any of our tests is *unsolvable* by multimodal transformer architectures. These shortcomings can be addressed through engineering solutions. For instance, one could implement adapters (Houlsby et al., 2019) specifically designed for additional visual downstream tasks (Hyeon-Woo et al., 2024), utilize specialized training with datasets tailored to capture the low- and mid-level properties of images (Bordes et al., 2024; Zanella & Ayed, 2024; Zhang et al., 2022), fine-tune the text decoder to better interpret visual representations (Huang et al., 2025), or employ an alternative visual encoder (Jiang et al., 2024; Jiao et al., 2024), which have been shown to capture fine-grained visual features (Caron et al., 2021; Oquab et al., 2024). Many of the match-to-sample tasks, for example comparing line lengths, could be completed by writing code to perform simple image calculations. While these strategies can effectively enhance the models' capabilities in these domains, they diverge from human-like cognition. In this study we prioritized using validated neuropsychological tests, which involved sacrificing some control over task design. One limitation arising from this is that the grouping of tests into task types (naming, match-to-sample, same/different, and yes/no) was not orthogonal to their grouping into visual stages or processes. For example, naming tasks appeared predominantly in high-level tests, which the models tended to perform better on. Having evenly distributed task types across groupings would provide a more balanced and controlled assessment. Using existing tests also meant that we had limited control over the number of trials. For example, the L-POST battery was designed for quick evaluation of mid-level impairments in patients and uses only five trials in each of its tests. These imbalanced trial numbers across tests mean that we have more precise estimates of model performance for some visual processes than others. We chose to employ a fixed set of prompt questions throughout (one for each task type), rather than tailoring prompts to maximise performance on each test. While this approach aligns with the clinical standards and reduces “researcher degrees of freedom” (Wicherts et al., 2016), it is worth acknowledging that changes in prompts may lead to differing task performance (He et al., 2024; Octavia & Cleti, 2024; Vatsal & Dubey, 2024).

Because our goal was to assess the most widely-used commercial state-of-the-art VLMs, here we test only closed-source models, which restricts insight into their mechanisms. Future research should explore open-source models (e.g., LLaVA, Qwen-VL, and DeepSeek-VL) for greater explainability of the model’s behaviour. Techniques like saliency maps (Jiang et al., 2024; Petsiuk et al., 2018) would allow researchers to probe the representation of images and text within models during the task. With the recent emergence of chain-of-thought (CoT) visual-language models (OpenAI, 2024; Shao et al., 2024), probing into a model’s reasoning may help reveal its response strategies and errors (although Lindsey et al. (2025) argued that CoTs are not always indicative of the internal processes or entirely faithful).

Extending the framework of using systematic assessments in human-model benchmarking of vision (Binz et al., 2024; Sheybani et al., 2024), future research could further test VLMs on basic visual assessments, including visual acuity, visual field sensitivity (similar but non-clinical work previously approached by Zhang et al., 2024), colour, and crowding, and other cortical visual assessments like the Visual Object and Space Perception batteries (VOSP) and the Cortical Vision Screening Test (CORVIST). Further testing into the domain of visual illusions (Ward, 2019; Zhang & Yoshida, 2024; Zhang et al., 2023) may also reveal interesting biases inherent in the models.

The widespread visual deficits in current VLMs demonstrate an example of Moravec’s Paradox within perception: a visual system does not need to be good at tasks that appear “basic” or “simple” to humans, in order to be good at “complex” tasks like object recognition. The way the human visual system has evolved and developed may make it especially suited to low- and mid-level visual judgements, in a way that artificial vision need not be.

## Method

### *Clinical Vision Tests*

We used 29 tests sourced from five clinical batteries. Full details of each test are provided in the Supplementary Methods. Low-level processes such as judgements of length, size, orientation, and position were assessed with the Birmingham Object Recognition Battery (BORB) (Humphreys & Riddoch, 1993). BORB also offers mid- and high-level visual assessments, such as identifying overlapping letters and naming objects.

Mid-level assessments offered by such classical clinical tests are often confounded with high-level processes as the stimuli largely rely on recognizable and nameable figures (Torfs et al., 2014). For this reason, we incorporated additional tools specifically targeting mid-level perceptual processes. The Leuven Perceptual Organization Test (L-POST) (Torfs et al., 2014) and the Leuven Embedded Figure Test (L-EFT) (de-Wit et al., 2017) were designed to assess mid-level perceptual organization such as grouping and figure-ground segregation without requiring semantic knowledge. L-POST offered over 15 tests, with two being object recognition-based tests acting as ecological tests to reflect mid-level challenges in real-world settings.

While BORB's high-level assessments required recognition of both objects and animals, we further expanded our testing into the face domain with the Dartmouth Face Perception Test (DFPT) (Dalrymple et al., 2017). We opted for DFPT over the commonly used Cambridge Face Memory Test (Duchaine & Nakayama, 2006), as the latter includes time constraints and a memory component that is unsuitable for VLMs. Despite DFPT being designed for young children, the match-to-sample task design employed was consistent with other tests we used. Additionally, we incorporated the Hooper Visual Organization Test (HVOT) to assess visuospatial perception where VLMs were required to identify objects based on their cutout fragments (Hooper, 1983).

### *Psychological Vision Tests*

We supplemented the clinical tests with 22 psychological vision tests, all generated from the MindSet toolbox. *MindSet: Vision* provides image datasets (or code to generate them) based on previous psychological findings for assessing DNNs (Biscione et al., 2024). Full test details are in the Supplementary Methods. Briefly, these datasets evaluate various visual abilities:

Weber's Law assessed length judgement, while Amodal Completion and Decomposition tested perceptual grouping. Non-accidental Properties vs. Metric Properties (NAP vs. MP) datasets examined sensitivity to 2D and 3D shape features. A small dataset was also included to probe sensitivity to relational and coordinate changes in object parts.

MindSet also included various object recognition datasets based of a subset of ImageNet categories, presented them as line drawings, silhouettes, and in modified forms (e.g., dotted contours, segmented outlines, and textures). Other datasets assessed sensitivity to global form modification and robustness to 2D transformations and 3D viewpoint shifts.

*Stimulus Preparation and Pre-processing*

We acquired neuropsychological tests in either physical or digital form. We obtained BORB as an eBook. We accessed L-POST and DFPT stimuli online with permission of the authors, while L-EFT was accessed at its public FigShare repository. A physical copy of the HVOT was obtained from the School of Psychology's clinical test library at the University of Auckland.

We obtained L-POST, L-EFT, and DFPT as sets of individual stimuli images, while BORB and HVOT as a digital and physical testing booklet. To create stimuli images, we scanned the physical copy of HVOT and then digitally cropped stimuli images, keeping the stimuli-to-border ratio approximately the same as the original presentation. We applied the same cropping process to BORB tests #7 to #14 to get individual stimuli images. BORB tests #1 and #9 were omitted due to them being drawing tasks, and the remaining were re-rendered for higher resolution. Images from BORB and HVOT were converted to grayscale to minimize color cast and glare.

We used source code from the MindSet: Vision toolbox to generate psychological datasets. We largely kept source code parameters as default. Unlike clinical tests with predetermined testing formats, we used stimuli images from the MindSet to form psychological tests based on a requirement that the testing format must be consistent with those used in the clinical tests. We reasoned that these test formats require low cognitive loads for humans (Kingdom & Prins, 2016) and are also adequately simple for model testing. For this reason, we omitted tests that do not fit these criteria, as task complexity could be confounded with perceptual performances. See Supplementary Methods for detailed stimuli preparation and testing.

All stimulus images were resized to 512x512 pixels (px) to control for the size and processing resolution.

*Model Evaluation*

We evaluated three state-of-the-art VLMs at the time of testing (August-November 2024): *GPT4o (2024-05-13)*, *Claude-3.5 Sonnet (new)*, and *Gemini-1.5 pro*. These models have demonstrated outstanding performance across various VLM benchmarks, showcasing their capabilities across multiple visual reasoning tests (Duan et al., 2024).

We provided the models with both instructions and images via the API, recording outputs for analysis. For clinical tests, we formatted our instruction prompts as closely as possible to the original tests to align with real-world clinical testing conditions (See test instructions in Supplementary Methods). We included additional statements instructing the model to provide a definitive final answer, even if it was uncertain. This approach mimics human clinical testing scenarios, where participants may be encouraged to use guesses or intuition to complete tasks when not confident of the answer, which can provide valuable insights into the difficulties. For psychological tests, we formatted the test instructions similarly to those used in clinical tests.

We tested the models in one of the four task formats: a match-to-sample task, a same-different task, a yes/no task, and a naming task. Each trial was submitted as a separate API request to prevent carryover context from one trial to the next. In match-to-sample, models selected the most similar image from two or three options. We clearly prompted the image label prior to each prompted image (e.g., “Target image”, “Option 1”, and “Option 2”). In same/different task, models judged whether two images differed in a specific characteristic. In yes/no task, the models judged whether a statement about a particular image property was true (e.g., whether two lines are parallel). In naming task, models named the object in an image. We recorded and scored the models' responses based on the clinical criteria for neuropsychological assessments and on previous psychological findings for psychological tests. Non-definitive answers were marked as incorrect for the trial.

*Online Behavioural Experiment*

We conducted an online behavioural experiment to collect human normative data for those tests for which human norms were not already provided. These were the psychological tests generated from MindSet, and four BORB tests that involved naming overlapping or non-overlapping triplets of shapes or letters, all of which were re-rendered stimuli. We included the BORB triplet tests because the original normative data was only available as response time. The experiment was created with Psychopy (Peirce et al., 2019) and hosted on Pavlovia (https://pavlovia.org/). This led to 27 tests (23 MindSet, one of which was later excluded in the analysis due to having only two trials; 4 BORB). We divided those into two sets of tests, completed in separate experimental sessions, each taking approximately 50 minutes (12 tests in Set 1 and 15 in Set 2).

Participants: We collected data from 60 participants (27 female; ages: 18-63 years). Participants were recruited via Prolific (https://app.prolific.com/) and were reimbursed the equivalent of 9 GBP per hour of participation. All participants reported normal or corrected-to-normal visual acuity with no other ocular pathologies, reported having English as their primary language, and lived in either the United Kingdom or the United States of America. The study procedures were approved by the University of Auckland Human Participants Ethics Committee (UAHPEC26388). Of the full sample, 56 participants completed both experimental sessions. We excluded 3 participants from each test set due to failing their corresponding attention check more than two out of three times. We further excluded 5 participants from Set 2 due to non-compliance in multiple tests (see Supplementary Methods for further details).

Procedure: Prior to the experiment, participants were required to read the participant information sheet and provide consent. The two experimental sessions had an identical format. Participants were first presented with three practice trials designed to familiarize them with the three task types (only three task types since the yes/no task did not appear among the experimental tests). In the match-to-sample task, we displayed three images on the screen, with the target stimulus on top and two image options at the bottom. The participants were asked to respond by clicking an image option. In the same-different task, we presented two images side-by-side, requiring the participant to press either ‘d’ or ‘s’ on the keyboard to indicate their answer. Naming tasks involved one target stimulus being displayed with a response box. Unlike the experimental blocks, the feedback was provided after each practice trial. All stimuli images were presented at 256x256 px resolution on a uniform grey background.

There were three experimental blocks. Each block contained one or more tests of the same task type. The block order and trial order within a block were randomized. We presented the participants with an attentional check question at the end of each block. The experiments were self-paced, with an 800 ms interval between one trial ending and the next being presented.

*Data and Statistical Analyses*

We categorized each test two ways 1) at a coarse level, according to which visual processing stage the test mainly assessed, and 2) for a finer-grain distinction, which specific perceptual abilities it primarily tested. In order to reach a consensus about these labels, our categorization was informed by an informal survey we conducted to canvas 11 professional vision scientists (excluding the authors). We asked them which coarse and fine-grained subgroup or subgroups they felt best described each test from among a list we provided.

As a result, we categorized the test into 1) three coarse groups: low-, mid-, and high-level visual tests and 2) nine finer-grained groupings: simple image element judgements, figure judgements, occlusion, and overlap, property biases, robustness across different image cues, robustness against different image configurations, recognition in visually straightforward circumstances, view invariance, and semantic association and categorization.

For each test, we compared the models' performance against human performance (measured in percentage). Human performances were obtained from the clinical normative data and behavioural data of our online experiment. We normalized the models' performance relative to humans' performance. Here, the relative performance of zero indicates that the model and human performed equally well in a test, whereas negative or positive values indicate that the model performance was worse or better than humans, respectively.

To test for overall differences in performance between the three models, we ran a model-label shuffling permutation analysis. For each pair of models, we calculated the mean difference between models, across all tests, and compared this observed difference to a null distribution of differences simulated by randomly shuffling the model labels for each test 10,000 times. Our models have overall similar accuracy levels (see Results). However, it is possible that they show distinct profiles of strengths and weaknesses across tests. To assess this, we calculated the Spearman correlation between each pair of models, subsequently showing high similarity across tests between our models. As a result, we averaged performances over all three models before comparing performance between different perceptual stages.

We tested for differences between tests grouped by us as tapping “low”, “middle”, or “high” level vision, and we ran a stage-label shuffling permutation analysis. For each pair of stages, we calculated the mean difference between stages, across all tests belonging to that stage. We then compared this observed difference to a null distribution simulated by randomly shuffling which tests were assigned to which stage (keeping the number of tests within each stage fixed), 10,000 times. We repeated the same analysis testing the differences between tests grouped by amongst our nine finer-grained test categories. We corrected for multiple comparisons using Bonferroni's correction on all permutation analyses.

## Contributions

**GT:** conceptualisation, data curation, formal analysis, investigation, methodology, software, visualisation, writing—original draft preparation, writing—review & editing; **KS:** conceptualisation, methodology, supervision, writing—review & editing.

## Acknowledgements

We thank Molly Walker for assistance in data processing, and Sam Schwarzkopf, Johannes Mehrer and Sarah Weigelt for discussions and feedback. KS was supported by a Marsden Fast Start Grant from the Royal Society of New Zealand (MFP-UOA2109).

## Supplementary Methods

### Stimulus Preparation and Testing Procedure

*Birmingham Object Recognition Test (BORB)*

We accessed BORB digitally online. BORB comprised 12 subtests, two of which were omitted. Subtests one and nine were drawing tasks not suitable for testing VLMs. The remaining BORB subtests were rerendered or preprocessed prior to model testing.

We re-rendered five BORB subtests to improve the resolution of the stimuli. For each trial of the length match task (subtest two), we rendered two stimuli images, each with a black horizontal line centered on the 512 x 512 canvas. Rather than converting the degree of visual angle of the stimuli length (cm), we replicated the stimuli configurations in the pixel space. With this, the relative size between lines remained proportionate in the same manner as the original stimuli. We rendered 30 trials (60 images) according to the original test. We repeated the same process with the size match task (subtest three) with 30 pairs, identical to the original. We tested the VLM on both tasks by asking it to judge whether two images of a line and a circle are of the same or different length and size, respectively.

In the orientation matching task (subtest four), each trial contained a pair of lines. The lines may either be parallel or not. We generated stimuli images of line pairs with their orientations (in degrees) being identical to the original test. Each stimuli image thus consisted of two possibly parallel lines with lengths of 200 px positioned 100 px vertically apart from the center. We rendered a total of 30 trials (60 images). The task for the VLM was to judge whether the two lines within each image were parallel.

In the position of gap task (subtest five), each trial contained two stimuli images, each containing a circle centered on a canvas of size 200 px with 2 px width. The orientation of the gap in degrees, starting from the top, was identical to the original stimuli. The size of the gaps was 20 px across all stimuli. Here, we rendered 40 trials (80 images) to replicate the original test set. For each trial, we required the VLM to determine whether the gap was located in the same location.

The overlapping figure tasks (subtest six), contained three stimuli sets- overlapping letters, shapes, and line drawings. For each set, we created two stimuli conditions- controlled and overlapped. In the control condition, the letters and shapes were presented as non-overlapping triplets. Letters triplets stimuli consisted of three 100 px Times News Roman

Regular letters centered on the canvas. The letters were randomly sampled from 26 uppercase English alphabets. Meanwhile, shape triplets consisted of three shapes of the size 100 px centered on the canvas. As we wanted to keep the stimuli-to-border ratio similar to the original, we left a 50 px gap between each shape in the triplet. The shapes were sampled from six possible shapes- cross, plus, triangle, square, pentagon, and hexagon.

For both letters and shapes, we generated the exact overlapped counterparts of the triplets. For each trial of overlapping letters, one letter was anchored to the center of the canvas while the remaining were randomly positioned around the 40 px radius of the anchor. We created overlapped shapes with a similar configuration but with a 75 px radius. We optimized these radii parameters based on our visual inspection to ensure that the degree of overlap is close to the original stimuli.

On the other hand, we cropped the overlapping line drawings from the original stimuli booklet. The control condition of line drawings consisted of a pair of non-overlapping objects. The overlapping counterpart, meanwhile, consists of objects from the same line drawings set but without the same object pairing. Across all conditions of the overlapping tasks, we asked the VLM to name all items in the images. We recorded the responses and scored the model based on the original criteria. With letters and shapes stimuli, the model received one point if it correctly identified the entire triplets of letters and shapes. Meanwhile, the model received one point for each object correctly identified in the line drawing tasks.

The remaining BORB subtests involved the use of line-drawing pictures. For these we cropped to isolate individual stimuli images at the stimuli's border, keeping the stimuli-to-background ratio approximately the same as the original presentation. During the testing of similarity judgment tasks, including the minimal feature view task (subtest seven), foreshortened view task (subtest eight), item match task (subtest eleven), and associative match task (subtest twelve), we presented the VLM with a target image followed by two image options. We prompted the VLM with similar instructions to that of the original test, asking the model to pick an image option that corresponds to the target image.

The object decision task contained a mixture of normal animals and objects and their amalgamated counterparts. The tasks contain four sets of stimuli- two labeled as easy and two labeled as hard. The task for the model was to identify whether the animal or object presented in the image was real or unreal.

Additionally, there were two sets of picture-naming tasks. One short form contained 15 items, while the long form contained 76.

All BORB stimuli were in black and displayed on a uniform white background at the resolution of 512x512 px.

*Dartmouth Face Perception Test (DFPT)*

We requested access to individual stimuli images of the DFPT via Testable. These images were rescaled to a resolution of 512x512 px. We replicated the original testing format and trial sequence. Each trial presented a target face, followed by three faces options. Among the three options, one was a distractor (a face from a different identity), while the other two were morphed faces of varying degree between the distractor and the target identity. The task instruction was kept as closely to the original as possible. Thus we asked the VLM the following: "*Below is a test. During the test you will see four face images. The first one will be the reference/target face image. The following three face images will be the options. Your task is to pick which of the following three face images matches the target. The faces will all look very similar, and choose carefully. You're able to give only one answer. Please answer with 1, 2, or 3*". We recorded the VLM chosen image and scored based on the answer key retrieved through Testable.

*Hooper Visual Organization Test (VOT)*

We scanned a physical booklet of the VOT and cropped the images to isolate individual stimuli near the stimulus borders. This preserved the original stimuli-to-background ratio. The images were then converted to grayscale to reduce colour casting and glare. The model was evaluated using the original test instructions, asking it to name the object in each image. Performance was scored according to the original clinical criteria, with one point awarded for correct responses and zero for incorrect ones. Some items allowed for awarding half a point for partially correct answers (Hooper, 1983).

*Leuven Embedded Figure Test (L-EFT) & Leuven Perceptual Organization Test (L-POST)*

We obtained L-EFT stimuli images via FigShare https://figshare.com/articles/dataset/Leuven_Embedded_Figures_Test_Context_Shapes/3807894) and the L-POST from the authors. The individual stimuli images were rescaled to 512x512 px resolution. Both tests followed a similarity judgment format, where the mode had to choose one of the three image options that best matched the target image. The L-EFT consisted of 64 trials. We assessed the VLM on this test with the following prompt, "*Below is a test. During the test you will see four images. The first one will be the target/reference. The following three are the* *options. The line drawing in the target image is embedded in one of the three option images. Your task is to decide for which figure this is the case. There is always only one correct answer. The target figure always has the same size and rotation, so that one of the figures among the options contains exactly the figure in the target image. You must provide an answer. Answer by saying either first, second, or third. Even If you are unsure, you will still have to pick an option.*"

The original L-POST comprised 15 subtests, four of which contain motion. Since VLM cannot yet process GIF animations, we excluded those subtests. This left us with 55 trials (11 subtests). Although each subtest targeted different mid-level visual abilities, they all followed the same prompt: "*Below is a test. During the test you will see four images. The first one will be the target/reference. The following three are the options. The line drawing in the target image is embedded in one of the three option images. Your task is to decide for which figure this is the case. There is always only one correct answer. The target figure always has the same size and rotation so that one of the figures among the options contains exactly the figure in the target image. You must provide an answer. Answer by saying either first, second, or third. If you are unsure you will still have to pick an option.*" We recorded the VLM response and awarded one point for each correct response.

*MindSet: Vision*

Mindset: Vision provides datasets of various psychological tasks with corresponding code scripts for image regeneration. The scripts offer multiple configurable parameters, though in most cases, we only adjusted the image size, rescaling the output to 512x512 px. Unlike clinical vision tests, MindSet: Vision provided only datasets of images and recommended testing methods with traditional CNNs. Therefore, we designed our own testing formats for each dataset while ensuring consistency with the task type and instructions used in clinical vision tests.

We limited our testing format to one of the same-different judgment task, similarity judgment task, or naming task. We determined the format that best suited each dataset. In the same-different tasks, the model was asked whether two images shared the same properties (e.g., line length). In similarity judgment tasks, the model chose between two images that best resembled the target image. We often used the following prompt, "*Here's a task. I will present you with three pictures. The first one is the target/reference image. The following two are the options. Your task is to choose which of the options most closely resembles the target/reference* *image. Please provide one final answer even if you are unsure.*" In the naming task, the model was asked to name the object shown in the image.

Amodal Completion Dataset. We generated 15 sets of samples. Each set contains samples of the distractors (Unoccluded), occluded, and notched conditions. Two variants of each set were generated, with either the square occluding the circle or vice versa. We have a total of 90 stimuli images (15 sets x 2 variants x 3 conditions). From this, we formed 30 similarity judgment trials, where each trial contained an image from each condition. The occluded condition was always the target. The model was required to choose whether the notched or the distractor condition was more similar to the target. In this task, humans are expected to find the distractor more similar to the target than the notched condition (Biscione et al., 2024; Rensink & Enns, 1998). Therefore, full marks in this task reflect the model responding the same way as humans.

Decomposition Dataset. Decomposition datasets comprised two types of shapes- familiar and unfamiliar. By default, a set of familiar shapes includes a triangle, square, rectangle, heptagon, circle, and wedge. Unfamiliar shapes were blob-like objects. We generated the stimuli images using default parameters, from which we formed 30 similarity judgment trials. We split half of the trials for familiar and unfamiliar shapes. Each trial displayed three images, one of each condition- natural, unnatural, and no split condition. No split condition where two shapes abut one another was always the target. We recorded the responses on whether the model chose the natural split (where two shapes were placed apart) or the unnatural split (where two shapes abut but one shape were struck with a straight line) as more closely matches the target. In this task, humans are expected to find the natural split condition more similar to the target (no split condition) than the unnatural split condition (Biscione et al., 2024; Jacob et al., 2021). Therefore, full marks in this task would reflect the model responding the same way as humans.

Non-accidental Properties (NAP) and Metric Properties (MP) for 2D lines and 3D geons Datasets. MindSet provided fixed sets of 2D line segments and 3D geons to evaluate the model's sensitivity to image manipulation. The 2D line dataset consists of 26 distinct line segments, while the 3D geon dataset includes 22 unique objects. Each 2D and 3D stimuli were generated in three conditions: reference, NAP change, and MP change. To assess the VLM's sensitivity to these changes, we created 26 similarity judgment trials for 2D stimuli and 22 for 3D stimuli. In each trial, we presented the model with the reference image, followed by NAP and MP image options. We recorded whether the VLM chose the NAP or MP variant as the variant that matches better with the reference. In these two tasks, humans would find higher similarity between the reference objects and their MP variants (Kayaert et al., 2003; Kubilius et al., 2017). Therefore, full marks in this task would reflect the model responding the same way as humans.

Relational versus Coordinate Changes Dataset. We generated five similarity judgment trials based on a fixed set of stimuli in the datasets. Each trial comprised a base shape, a base shape with relational change to its parts, and a base shape with coordinate change to its parts. We recorded the model answer on whether it judged the relational or coordinate change as a better match to the base shape. In this task, humans would find higher similarity between the base shapes and their coordinate modification counterparts as opposed to their relational change (Biscione et al., 2024). Full marks in this task where would reflect the model responding the same way as humans.

Weber Law Dataset: Weber's Law dataset provided images of a white horizontal line on a black background, varying in both length (in pixels) and pixel intensity (maximum of 255). We configured the script to generate images with line lengths ranging from 20 to 420 pixels, in 20-pixel intervals, and at two intensity levels- low contrast (55 a.u.) and high contrast (255 a.u.). We created 20 same-different judgment trials for each contrast condition, resulting in a total of 40 trials. The line lengths for each trial were randomly paired, but the pairings were identical across both contrast conditions. During the test, we inputted the model with two images and recorded whether it correctly indicated if the two lines had the same length.

Line Drawings, Dotted Line Drawings, Segmented Images, and Silhouettes. MindSet provided a variety of datasets for testing DNN's object recognition across different image domains. These datasets comprised 36 ImageNet classes, with images drawn using solid white strokes, dotted and segmented outlines, or shaded with white fill on a uniform black background. We generated the datasets using default parameters, except the image resolution. We formed naming tasks with the number of trials determined by the generated outputs. Therefore there were 36 images, each of distinct classes for line drawings, dotted line drawings, and silhouette conditions. The segmented images dataset contained two complementary image segments that together combined to create the original line drawings. The segmented image dataset thus contained 72 total trials. We tested the VLM's ability to identify the objects in the image with the following prompt, "*Here's a task. Could you please name the item in this* *image?."* This prompt remained consistent with those used in clinical naming tasks. During scoring, since ImageNet labels include a mix of basic and subordinate-level categories, we awarded one point of the model correctly identified at least the basic-level category.

2D Transformations dataset. 2D transformations dataset applied affine transformations (rotation, translation, scale, and shear) to the line drawing dataset. We generated 5 samples for each of the 36 classes with default transformation parameters. We then randomly selected 2 samples of each class to form the naming task. We excluded one object class- *white wolf*- as the generated images were faulty, resulting in 70 trials. The VLM was tested on its ability to name the object in the image. We scored the model in the same manner as the line drawings and silhouette datasets.

Texturized blobs and silhouettes. MindSet provided texturized datasets of both familiar objects and texturized blobs. Familiar object datasets were derived from the 36 ImageNet classes used in line drawing tasks. The objects were texturized by masking the internal contour of the silhouette with either a pattern of repeated characters with random font size and rotation, or short line segments with a random length and slope. We tested the model's ability to name the texturized objects on both texture types, each containing 36 randomly sampled images of distinct classes.

Texturized blobs were generated using the same approach. The blobs were texturized with random characters. We tested the VLM's ability to recognize blobs under varying textures with a similarity judgment task. MindSet generated 10 unique blobs which we used to form 20 trials. Each blob shape was chosen twice as the target. Target blobs were always solid white silhouettes. During each trial, we presented the model with the target blob and two image options, one of which was the randomly selected texturized counterpart of the target, and the other was a randomly selected texturized blob of a different shape. We scored the model on whether it correctly identified the matching blob to the target.

Global change lines and silhouettes. MindSet provided two datasets to test the model's invariance to global changes in objects. There were two datasets corresponding to each image domains- line drawings and silhouettes, respectively. The line drawings dataset was based on the 36-class line drawings used in previous tasks, while the silhouettes dataset was derived from a 9-classes dataset (cite). We generated both datasets using default parameters. Each dataset contained three conditions (see Biscione et al., 2024 Figure 13)- the control, fragmented, and frankenstein. The control conditions were the original line drawings and silhouettes. The fragmented conditions were made up of the original images horizontally sliced into two halves, with the top being flipped and offset. The frankenstein condition was similar but without the offset, with the flipped half abutting the bottom half.

We therefore had six sets of images for naming tasks in total, each with 36 trials. In the line drawings dataset, each trail featured a unique class, while the silhouette dataset contained 36 trials drawn from 9 classes, each appearing four times. We generated 40 variants of each silhouette class and randomly sampled four per class to create the trials. The scoring procedure remained consistent with previous naming tasks.

Same different task. Stimuli in the same-different dataset involve images containing two shapes randomly placed on a black uniform canvas. The dataset contained 10 conditions, such as regular (geometric), irregular, straight lines, and segmented lines. Two shapes in the images were either the same or different. We generated five variations of the same and different stimuli for each condition (the remaining parameters were kept as default). We then randomly sampled three for the same-different tasks, resulting in 60 trials (3 variants x 2 same/different x 10 conditions). During testing, we presented the model with a single image each trial with the following instruction, "*Here's a task. There are two items randomly placed in the image. The two items can be either the same or different. Your task is to tell me whether the two items in the image are of the same or different. Please provide one final answer, even if you are unsure.*"

Viewpoint Invariance. MindSet built the viewpoint invariance dataset using the ETH-80 dataset. The dataset contained images of eight categories, each consisting of 10 object instances captured at viewpoints varied across four steps of elevation and 25 azimuth angles. We generated 32 similarity judgment trials. Each object categories were included four times, each with different object instances. During testing, we presented the model with the image of a target instance, followed by two image options. One of the options matches the instance of the target but viewed from a slightly different angle (within 90 degrees azimuth), while the other is a different instance in the same category viewed from the same viewpoint as the target instance. We prompted the model with the following instruction, "*Here's a task. I will present you with three pictures. The first one is the target/reference image. The following two are the options. One is an image of the exact same target viewed from a different angle, while the other is a different object. Your task is to choose which of the options is the same exact object as the target/reference image. Please provide one final answer, even if you are unsure.*" We recorded whether the VLM chose the image option with the same instance as the target to assess its viewpoint invariance.

Excluded Tasks and Datasets. Among MindSet's low to mid-level vision dataset, we excluded the Emergent Feature and Crowding dataset from our testing as we deemed those unsuitable for model testing. We further excluded the Depth Drawings dataset from our final analysis as the task only contained two samples. MindSet additionally offered various datasets for visual illusion. We did not include these as we aim to focus on fundamental visual processes.

We excluded BORB overlapped and paired line drawing tasks as the clinical normative data provided involved only response times. Unlike, BORB overlapped letters and shapes, we did not rerender the line drawings and included them in our human testing.

Amongst the MindSet high-level object recognition dataset, we removed trials associated with Hammer, Teapot, Toucan, and Penguin from datasets based on shaded images such as silhouettes and texturized datasets. The stimuli generated by the toolbox for these objects were incomplete. For instance, the texture was only applied to the head, not the handle of the hammer.

**Online behavioural study**

*Data preprocessing*

We first checked participants' compliance based on their scores across three attention checks in each experiment set. We excluded sets in which participants failed more than two times. We further checked unusually low scores on naming tasks (those with scores more than 2 standard deviations below the task mean) were manually checked and re-marked where necessary to include valid responses missed by automated scoring (e.g., due to spelling and formatting errors). Throughout this process, we excluded participants whose responses consistently showed non-compliance to task instructions and those who did not appropriately complete the experiment.

## Supplementary Table

| Name of task | Source battery | Name in source | Task type | Number of trials | Example trial | Visual ability | Visual stage |
|---|---|---|---|---|---|---|---|
| Gap position comparison | BORB* | “Position of gap” | Same / Different | 40 | 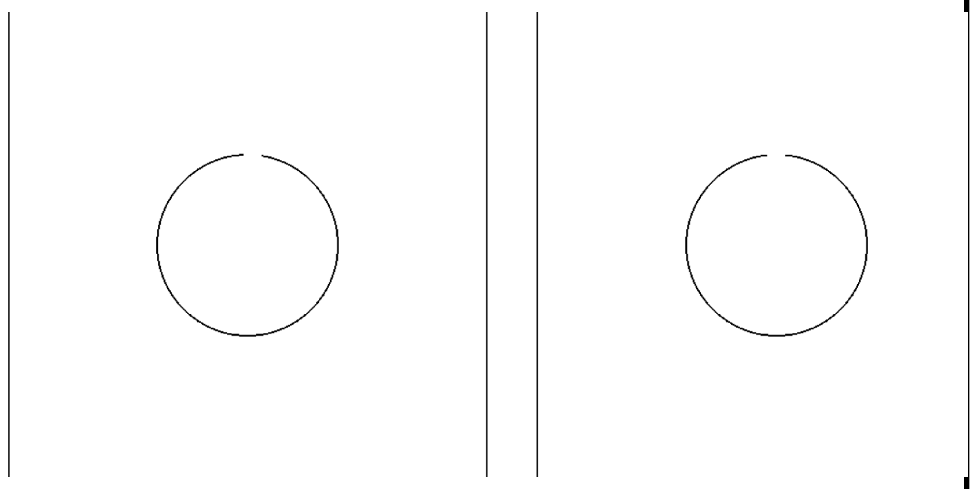 | Simple element judgement | Low-level |
| Line length comparison | BORB* | “Length match” | Same / Different | 30 | 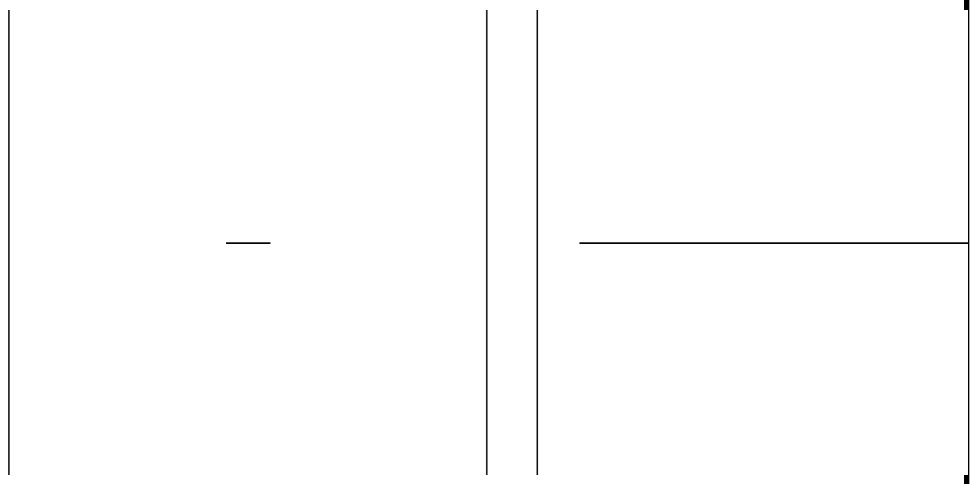 | | |
| 2D aspect ratio matching | L-POST | “Shape ratio discrimination” | Match-to - sample | 5 | 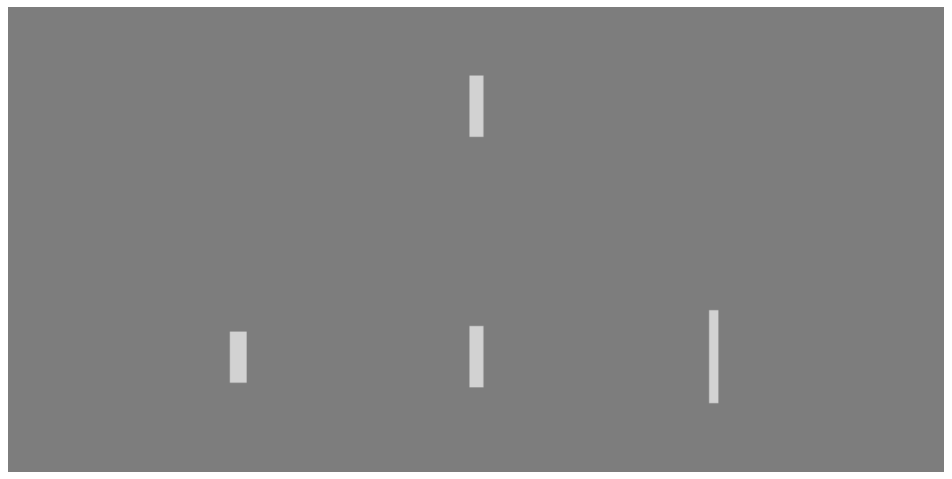 | | |
| Texture orientation matching | L-POST | “Dot Lattices” | Match-to - sample | 5 | 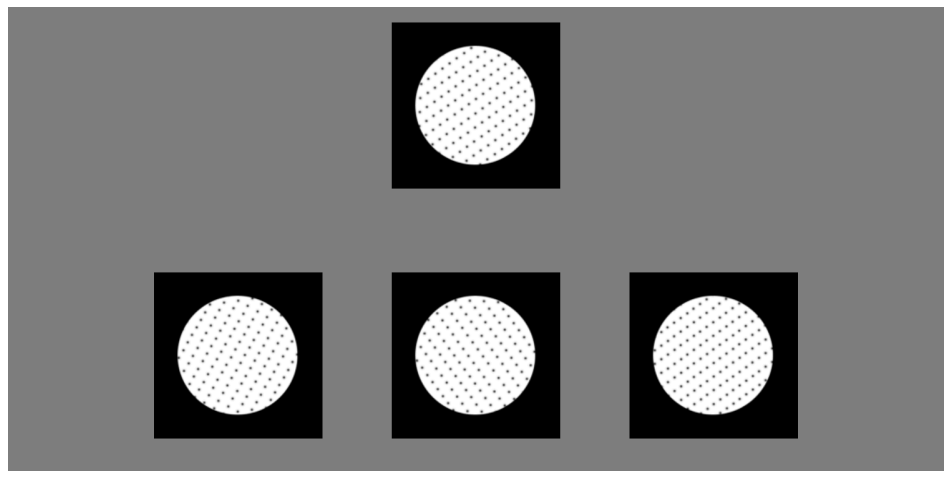 | Simple element judgement (cont’d) | Low-Level (cont’d) |
| Parallel lines judgement | BORB* | “Orientation match” | Yes/no | 30 | 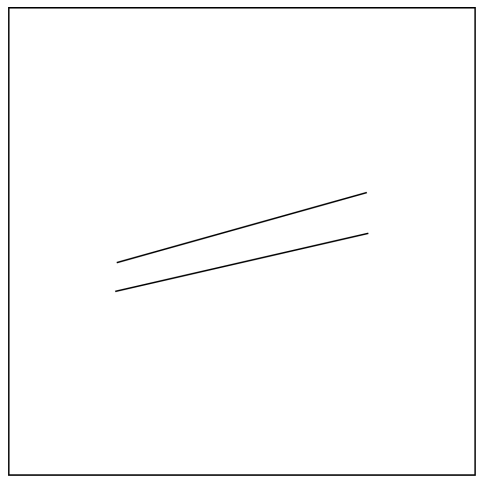 | | |
| Size comparison | BORB* | “Size match” | Same / Different | 30 | 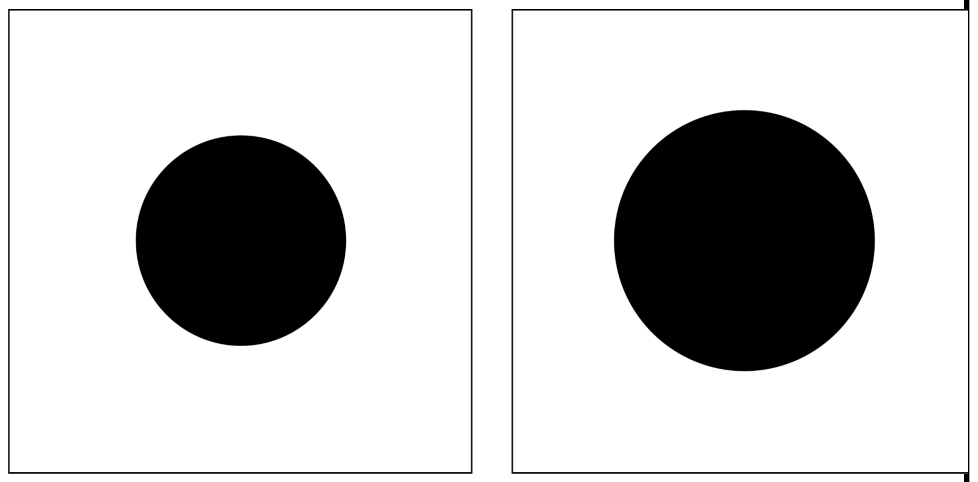 | | |
| Line length comparison | MindSet | “Weber’s law” | Same / Different | 40 | 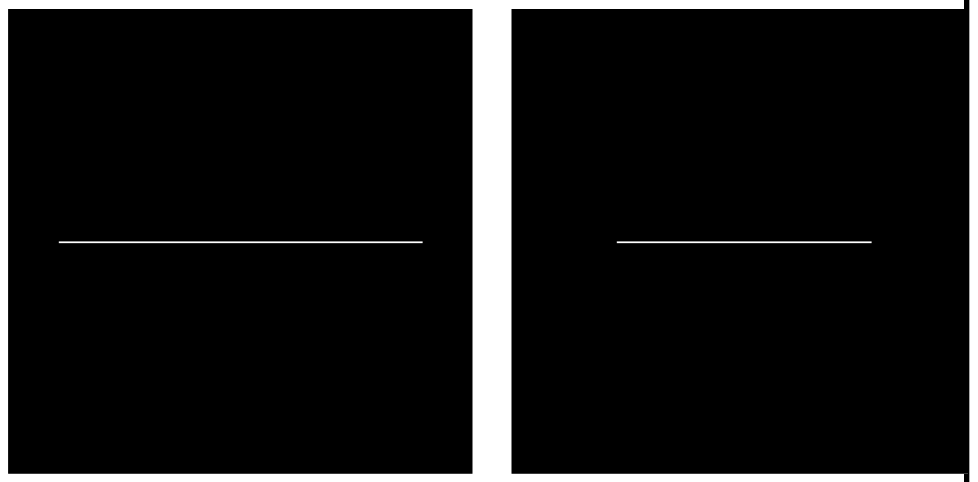 | | |

<table>
<tr><td>Embedded figure matching</td><td>L-EFT</td><td>“Embedded figure detection”</td><td>Match-to - sample</td><td>64</td><td>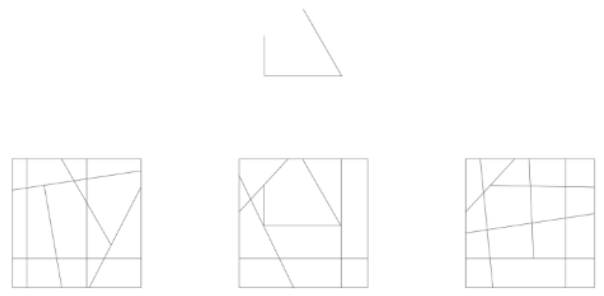</td><td rowspan="4">Figure judgement</td><td rowspan="4">Mid-level</td></tr>
<tr><td>Embedded figure matching</td><td>L-POST</td><td>“Embedded Figure Detection”</td><td>Match-to - sample</td><td>5</td><td>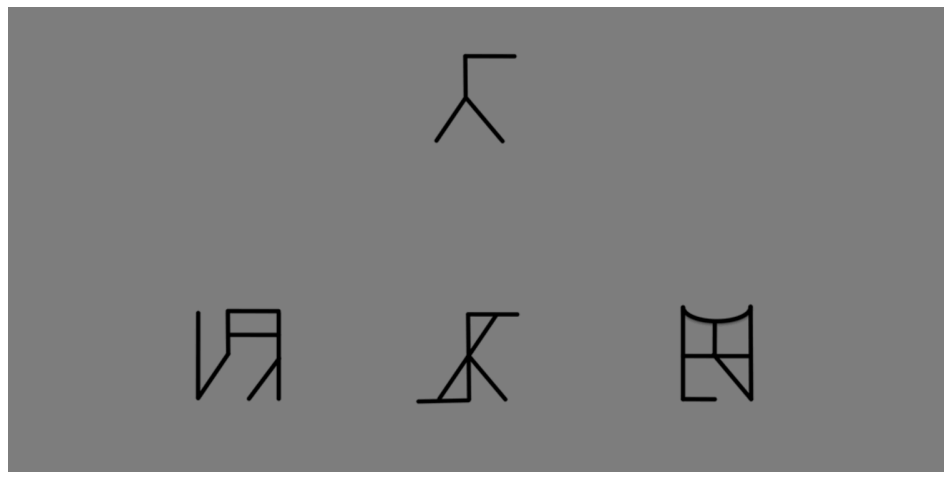</td></tr>
<tr><td>Missing part matching</td><td>L-POST</td><td>“Recognition of missing part”</td><td>Match-to - sample</td><td>5</td><td>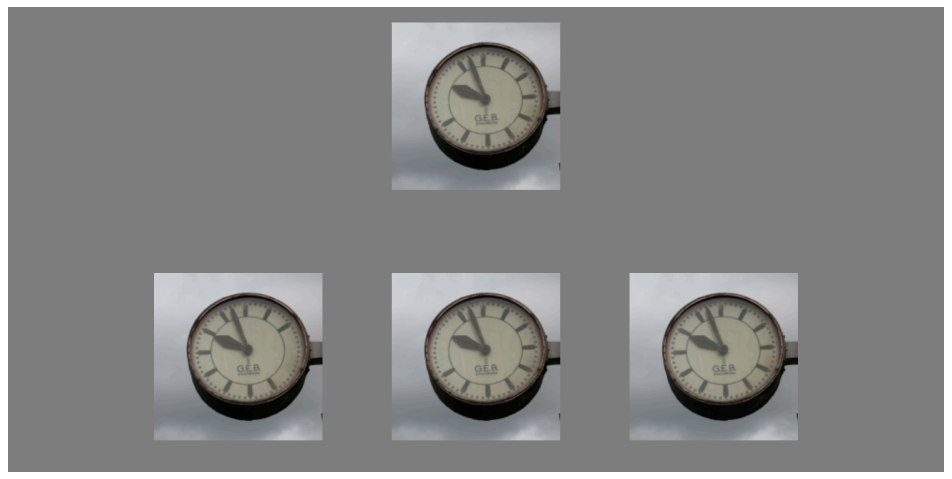</td></tr>
<tr><td>3D shapes matching</td><td>L-POST</td><td>“Fine shape discrimination”</td><td>Match-to - sample</td><td>5</td><td>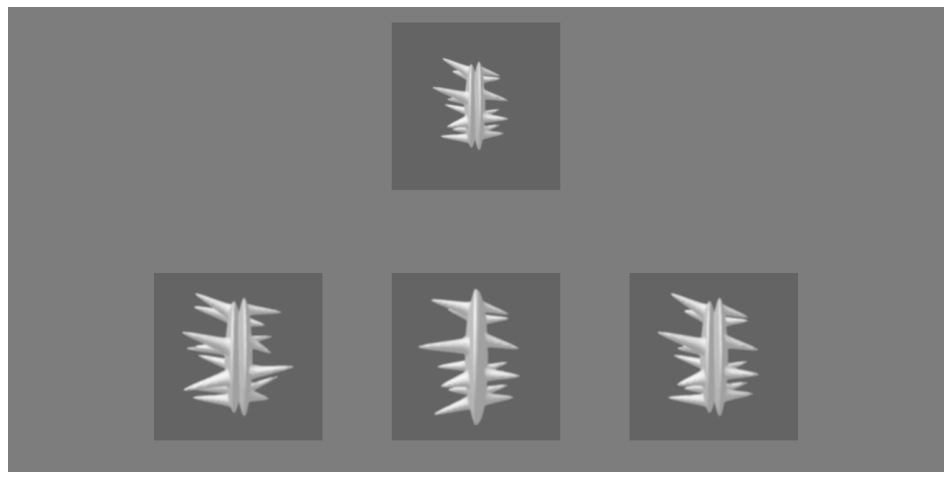</td></tr>
</table>

| | | | | | | | |
|---|---|---|---|---|---|---|---|
| Geometric elements comparison | MindSet | "Same different task" | Same / Different | 60 | 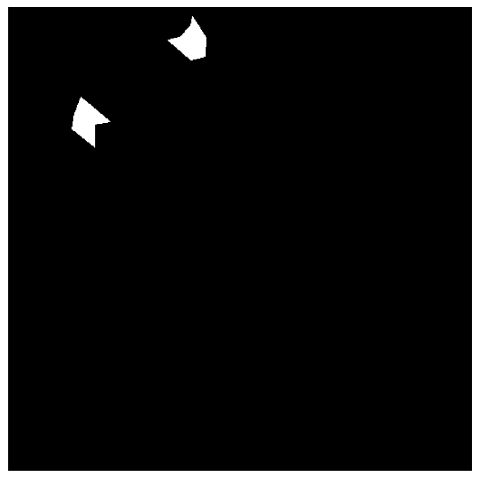 | Figure judgement (cont'd) | |
| Occluded shape matching | L-POST | "Figure-ground segmentation" | Match-to - sample | 5 | 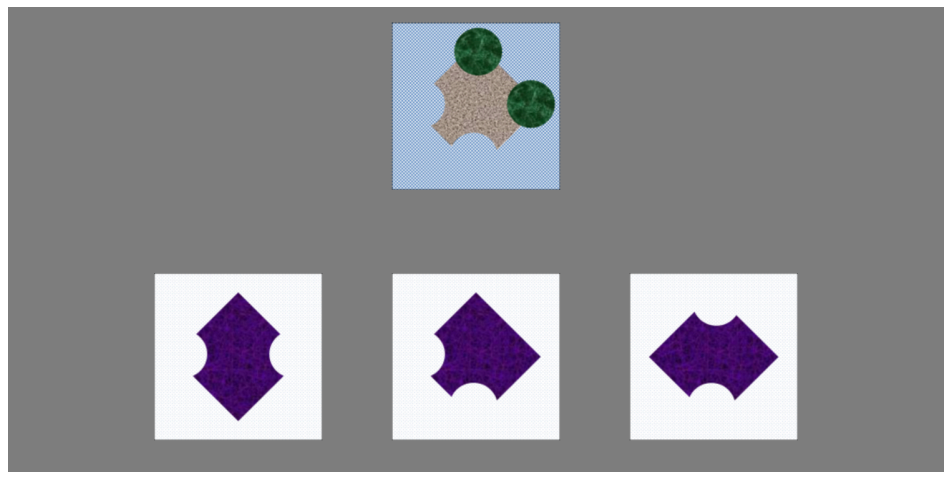 | | |
| Overlapping letters identification | BORB* | "Overlapping letters" | Naming | 36 | 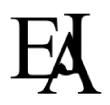 | Occlusion and overlap | Mid-level (cont'd) |
| Overlapping shapes identification | BORB* | "Overlapping shapes" | Naming | 36 | 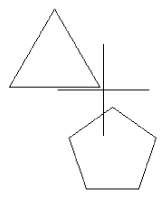 | | |
| Occluded shape matching | MindSet | "Amodal completion" | Match-to - sample | 30 | 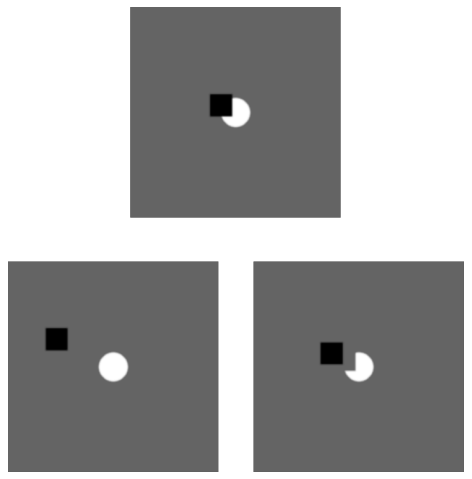 | Occlusion and overlap (cont'd) | Mid-level (cont'd) |
| 2D property change matching | MindSet | "Non-accidental property (NAP) vs metric property (MP): line segments" | Match-to - sample | 26 | 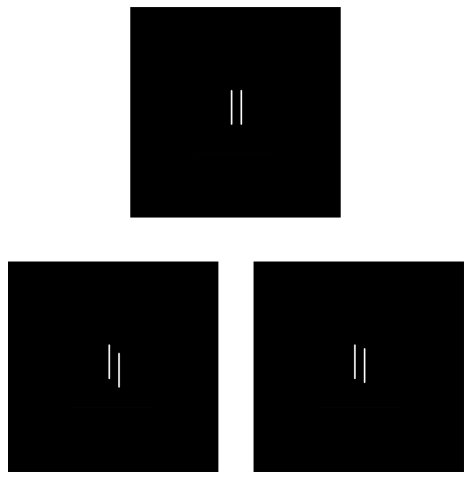 | Property biases | |
| Separated shape matching | MindSet | "De-composition" | Match-to - sample | 30 | 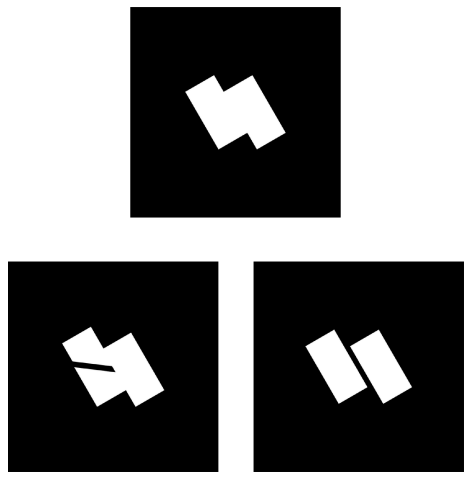 | | |
| Configuration change matching | MindSet | "Relational vs coordinate change" | Match-to - sample | 6 | 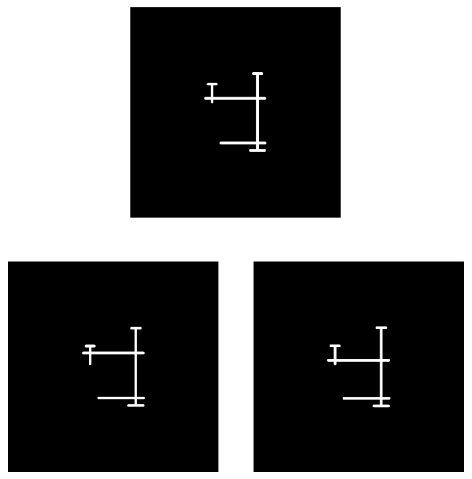 | | |
| 3D property change matching | MindSet | “Non-accidental property (NAP) vs metric property (MP): geons” | Match-to-sample | 22 | 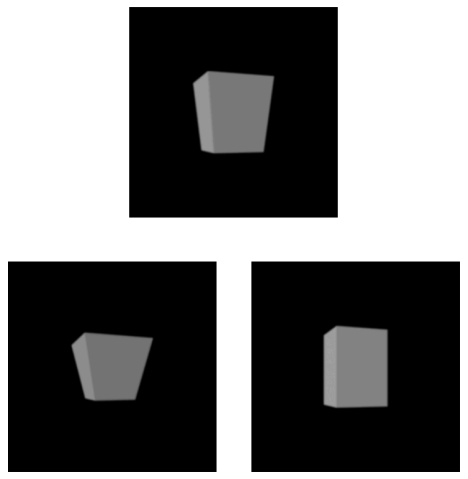 | Property biases (cont’d) | Mid-level (cont’d) |
| Shape matching (partial outline) | L-POST | “RFP fragmented outline” | Match-to - sample | 5 | 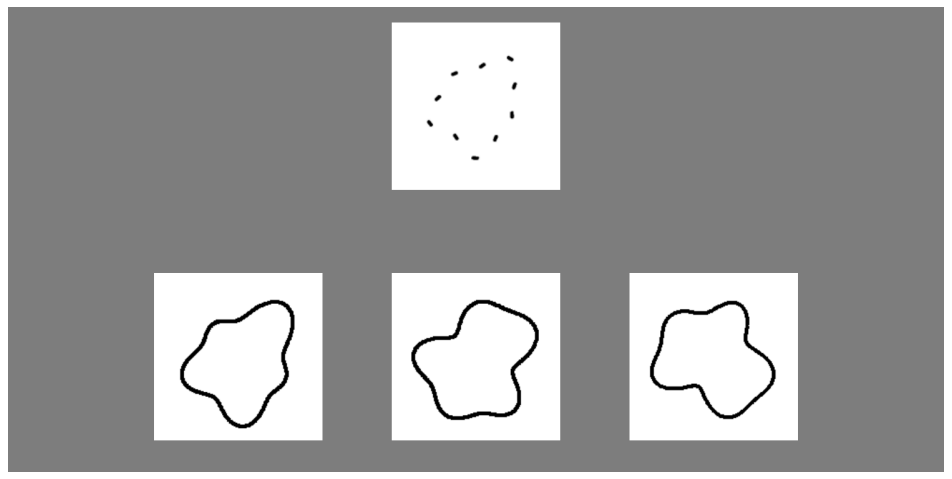 | Robustness across image cues | |
| Shape matching (hidden contour) | L-POST | “RFP contour integration” | Match-to - sample | 5 | 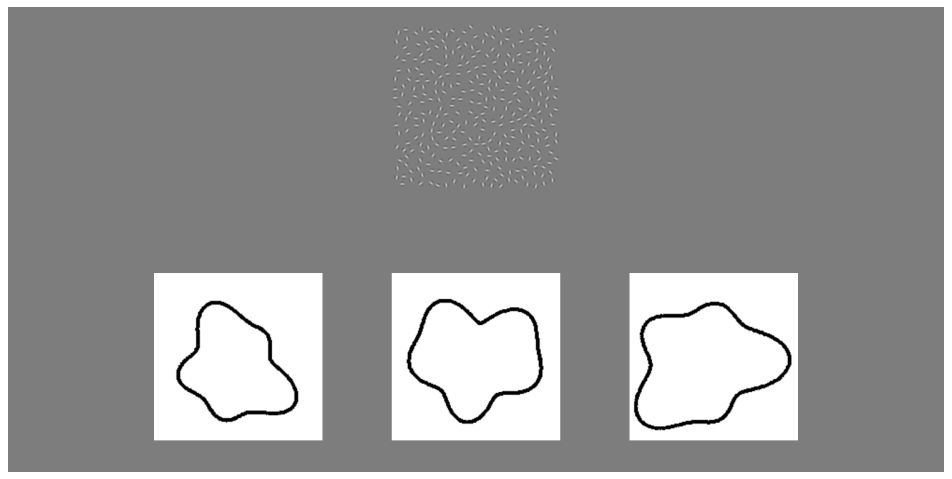 | | |
| Shape matching (texture-defined) | L-POST | “RFP texture surfaces” | Match-to - sample | 5 | 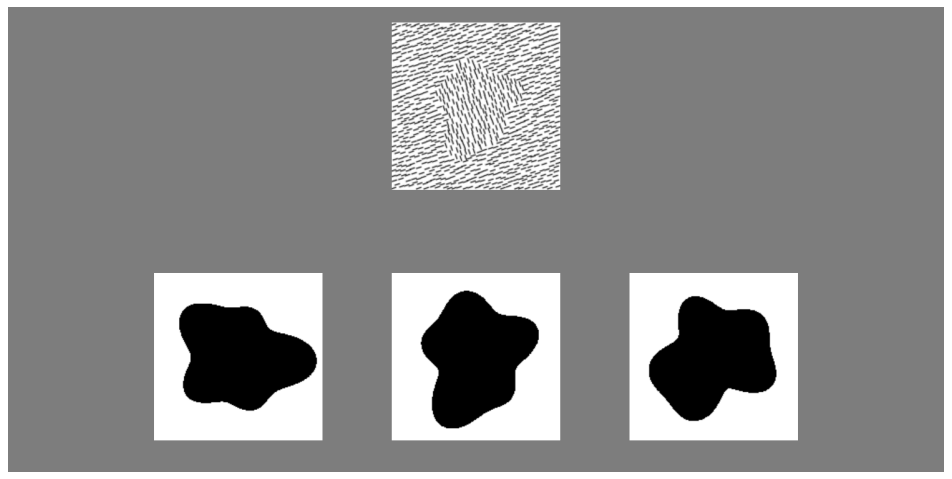 | | |

<table>
<tr><td>Shape-matching (texture-defined)</td><td>MindSet</td><td>“Texturized Blob”</td><td>Match-to - sample</td><td>20</td><td>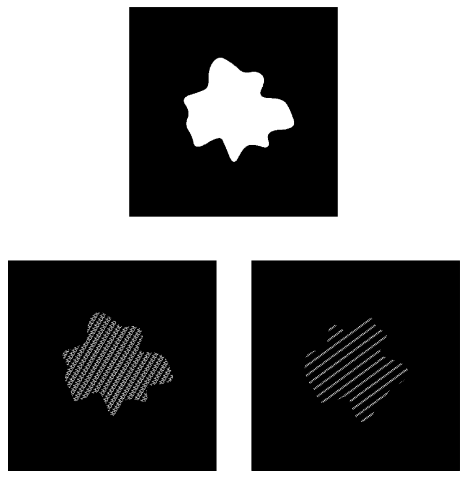</td><td rowspan="4">Robustness across image cues (cont’d)</td><td rowspan="4">Mid-level (cont’d)</td></tr>
<tr><td>Object naming (texture-defined chars)</td><td>MindSet</td><td>“Texturized line drawings”</td><td>Naming</td><td>32</td><td>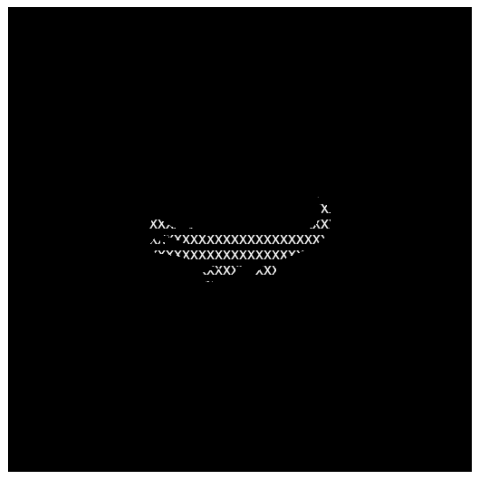</td></tr>
<tr><td>Object naming (texture-defined lines)</td><td>MindSet</td><td>“Texturized line drawings”</td><td>Naming</td><td>32</td><td>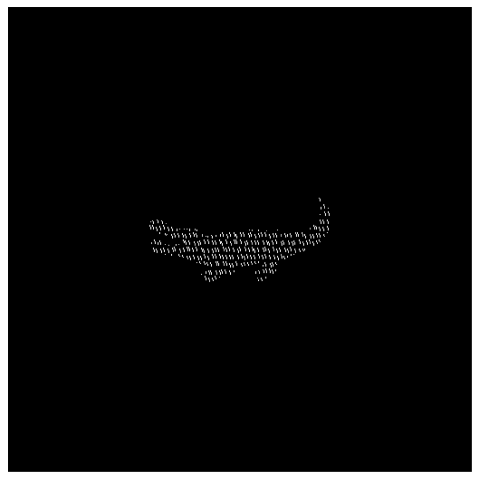</td></tr>
<tr><td>Object naming (partial outline)</td><td>MindSet</td><td>“Segmented image”</td><td>Naming</td><td>72</td><td>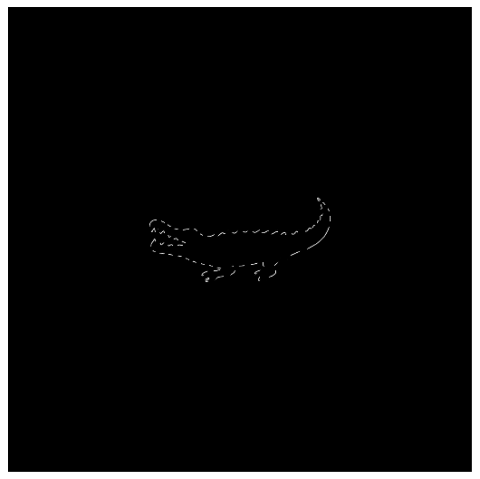</td></tr>
</table>

| | | | | | | | |
|---|---|---|---|---|---|---|---|
| Object naming (dotted outline) | MindSet | “Dotted line drawings” | Naming | 36 | 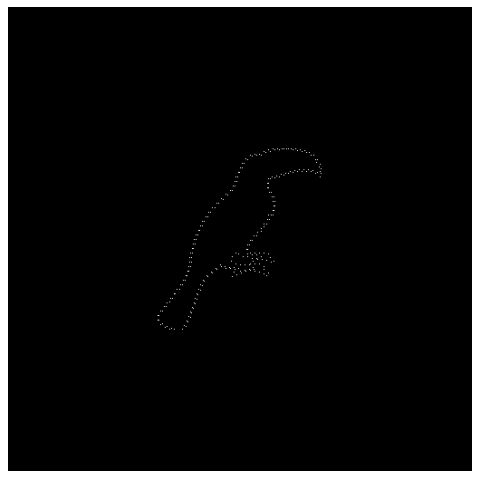 | Robustness across image cues (cont’d) | Mid-level (cont’d) |
| Segmented object identification | HVOT* | “Visual organization test” | Naming | 30 | 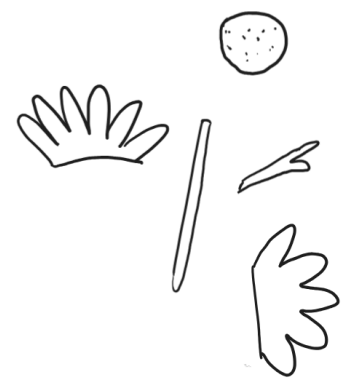 | Robustness across different configurations | |
| GC Object naming (outline fragmented) | MindSet | “Global change fragmented line drawings” | Naming | 36 | 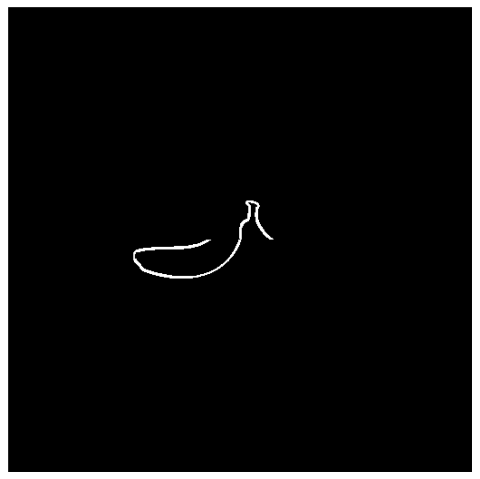 | | |
| GC Object naming (silhouette fragmented) | MindSet | “Global change fragmented silhouettes” | Naming | 36 | 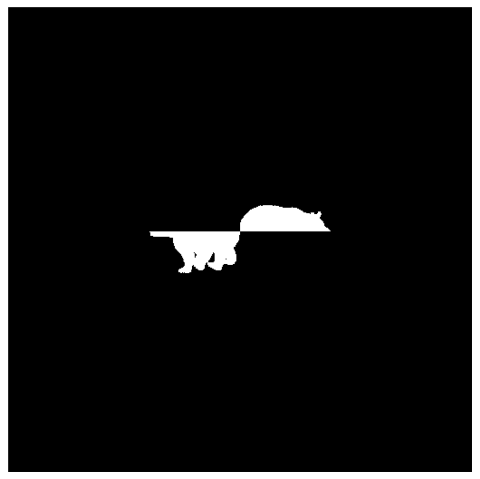 | | |
| GC Object naming (outline frankenstein) | MindSet | “Global change Frankenstein line drawings” | Naming | 36 | 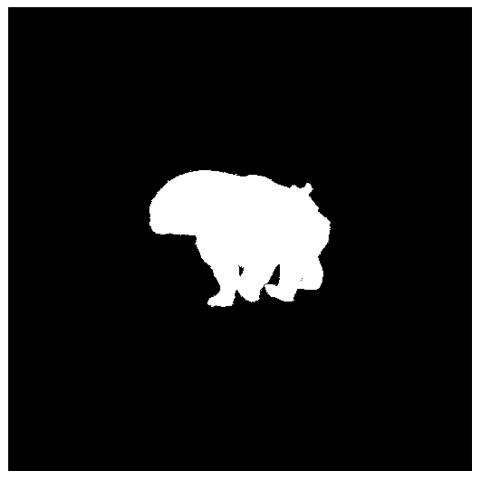 | Robustness across different configurations (cont’d) | Mid-level (cont’d) |
| Object naming (affine transform) | MindSet | “2D trans-formations” | Naming | 70 | 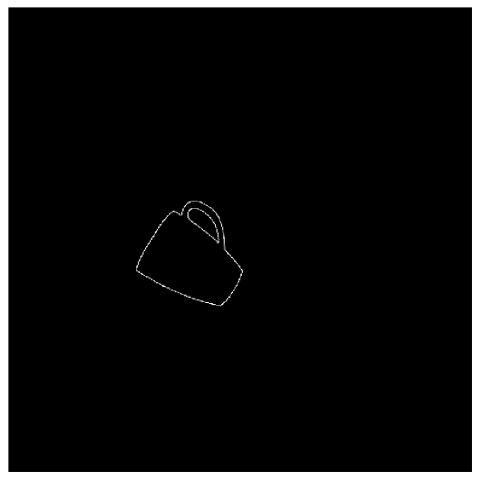 | | |
| GC Object naming (silhouette frankenstein) | MindSet | “Global change Frankenstein silhouettes” | Naming | 36 | 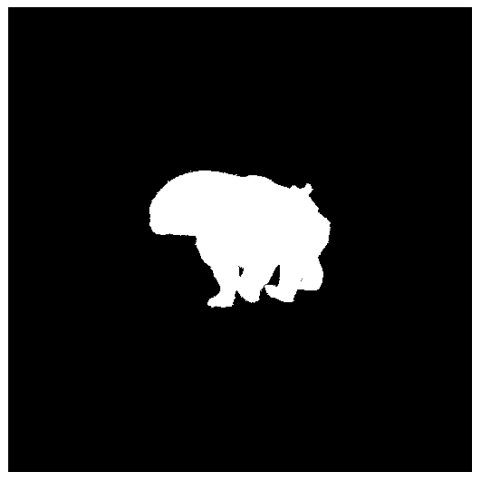 | | |
| Object naming (short) | BORB* | “Object naming (short)” | Naming | 15 | 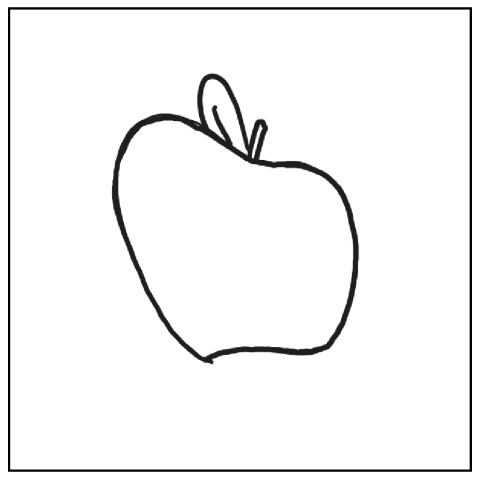 | Straight-forward visual recognition | High-Level |
| Object naming (long) | BORB* | “Object naming (long)” | Naming | 76 | 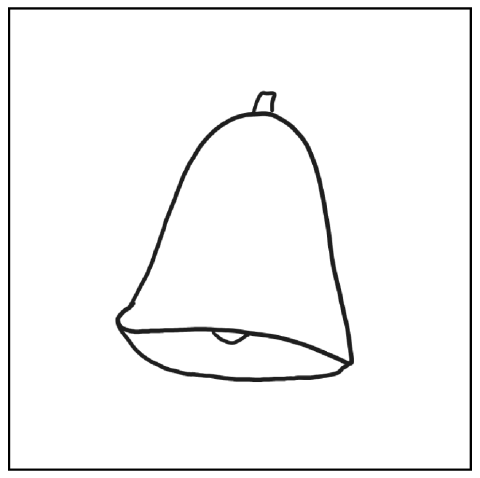 | | |
| Object naming in scene | L-POST | “Recognition of objects in a scene” | Naming | 5 | 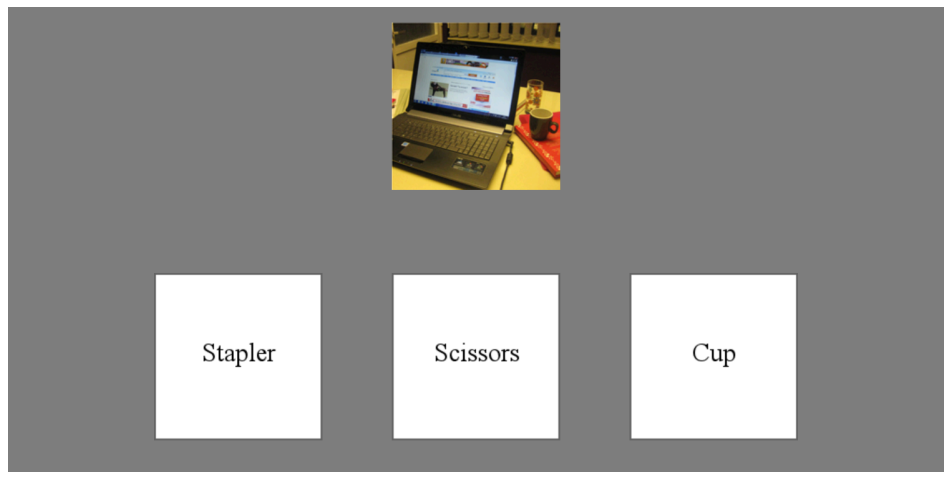 | | |
| Letters identification | BORB* | “Triplet letters” | Naming | 36 | EJA | Straight-forward visual recognition (cont’d) | High-Level (cont’d) |
| Shapes identification | BORB* | “Triplet shapes” | Naming | 36 | 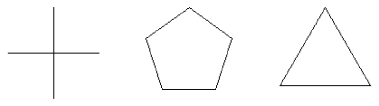 | | |
| Isolated object naming | L-POST | "Recognition of objects in isolation" | Naming | 5 | 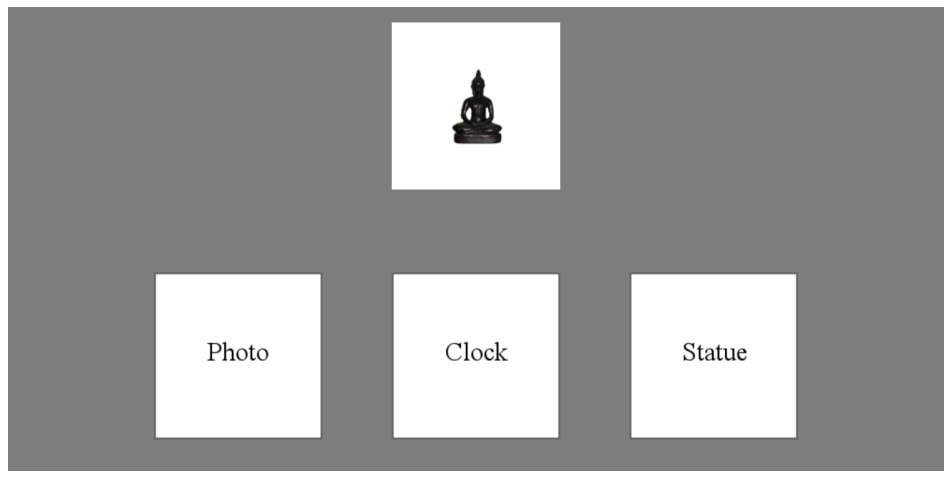 | Straight-forward visual recognition (cont'd) | High-Level (cont'd) |
| GC object naming (outline) | MindSet | "Global change original line drawings" | Naming | 36 | 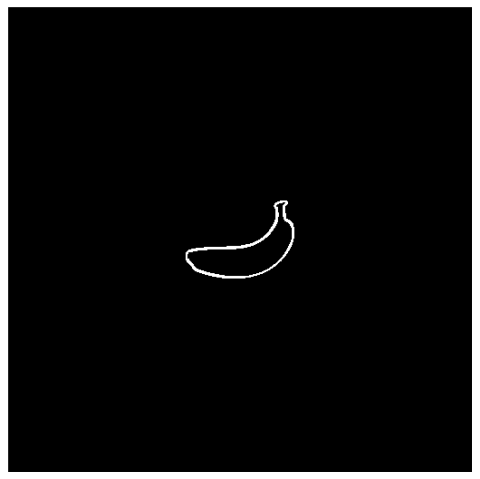 | | |
| Object naming (outline) | MindSet | "Line drawings" | Naming | 36 | 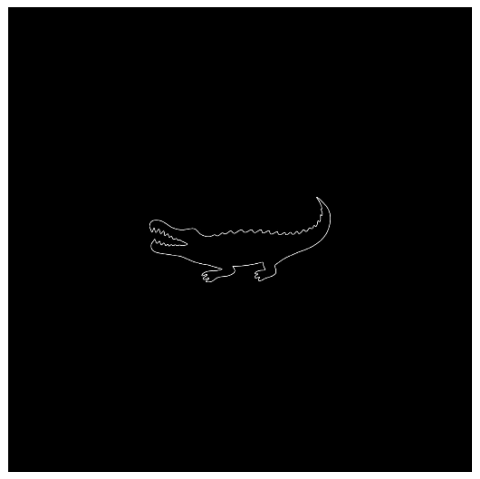 | | |
| GC object naming (silhouette) | MindSet | "Global change original silhouettes" | Naming | 36 | 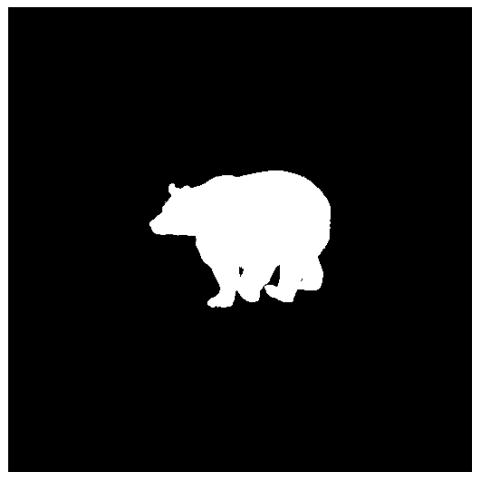 | | |
| Object naming (silhouette) | MindSet | “Silhouettes” | Naming | 32 | 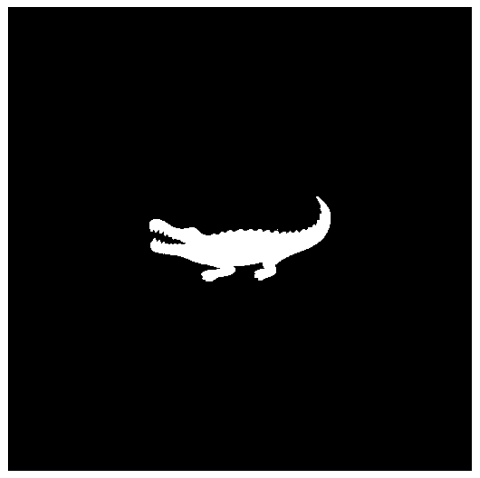 | Straight-forward visual recognition (cont’d) | High-Level (cont’d) |
| Object matching across views (minimal features) | BORB* | “Minimal feature view” | Match-to - sample | 25 | 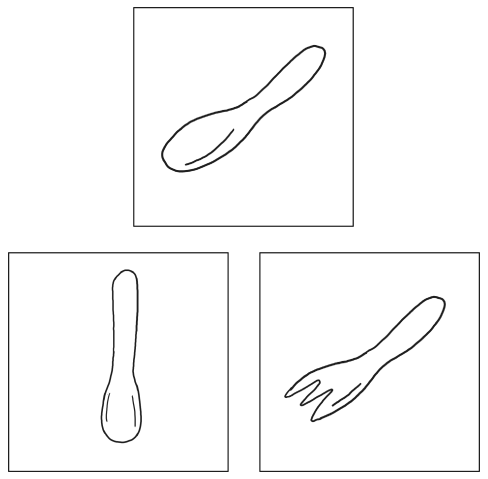 | View invariance | |
| Object matching across views (foreshortened) | BORB* | “Fore-shortened view” | Match-to - sample | 25 | 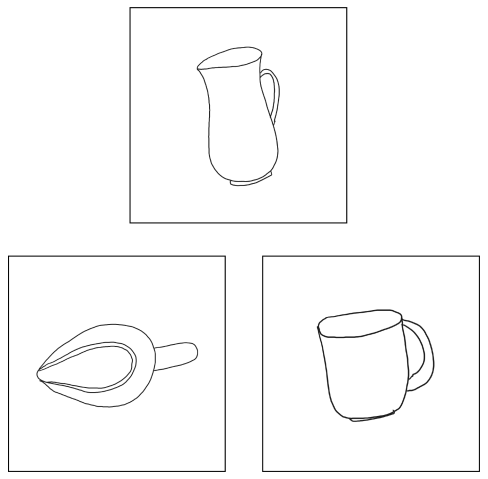 | | |
| Face matching | DFPT | “Face perception task” | Match-to - sample | 40 | 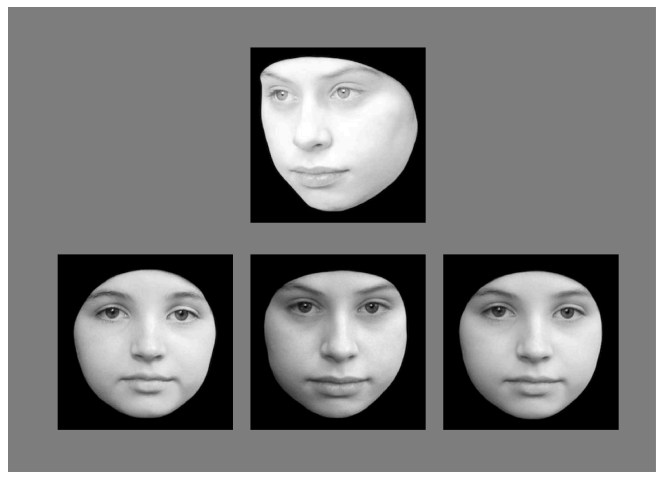 | | |
| Object drawings matching | BORB* | “Item match” | Match-to - sample | 32 | 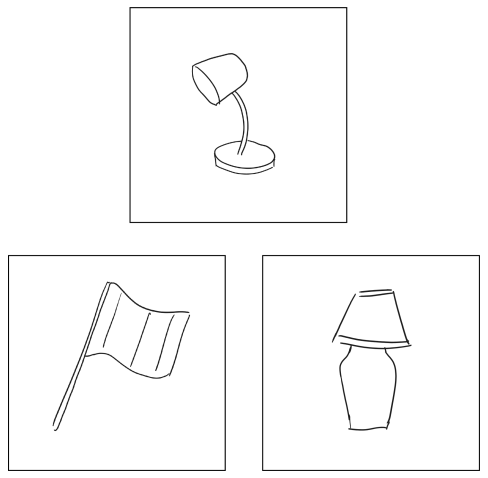 | View invariance (cont’d) | High-Level (cont’d) |
| Object matching across views | MindSet | “Viewpoint invariance ” | Match-to - sample | 32 | 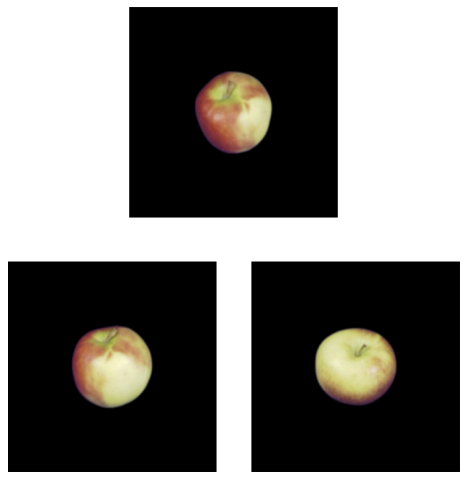 | | |
| Object associations matching | BORB* | “Associati ve match” | Match-to - sample | 30 | 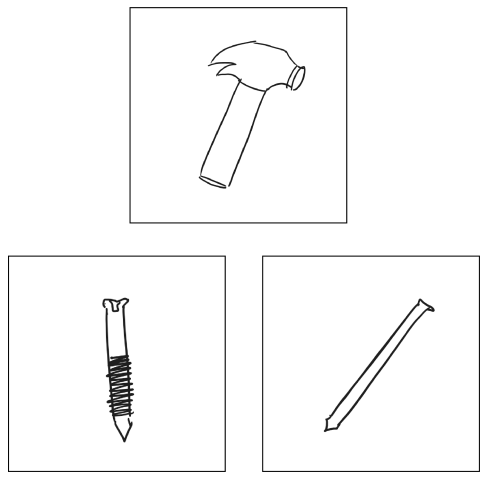 | Semantic associatio n and categoris ation | |
| Real/unreal object judgement | BORB* | “Object decision” | Match-to - sample | 128 | 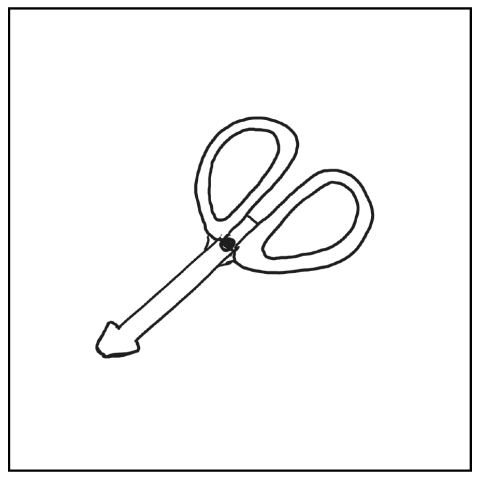 | | |

**Supplementary Table S1: Details and examples for each task.** *“Name of task”* matches that used in Figure 2 and throughout the text. *“Name in source”* refers to the name used to refer to the task in the source battery, if any. For tasks in open-source batteries, we show image(s) from one

example trial. For tasks in closed-source clinical batteries (BORB and HVOT, indicated with an asterisk *), we show a recreation of image(s) from one example trial. *"Visual ability"* and *"Visual stage"* were designated by us to broadly group similar tasks together.

## Supplementary Figure

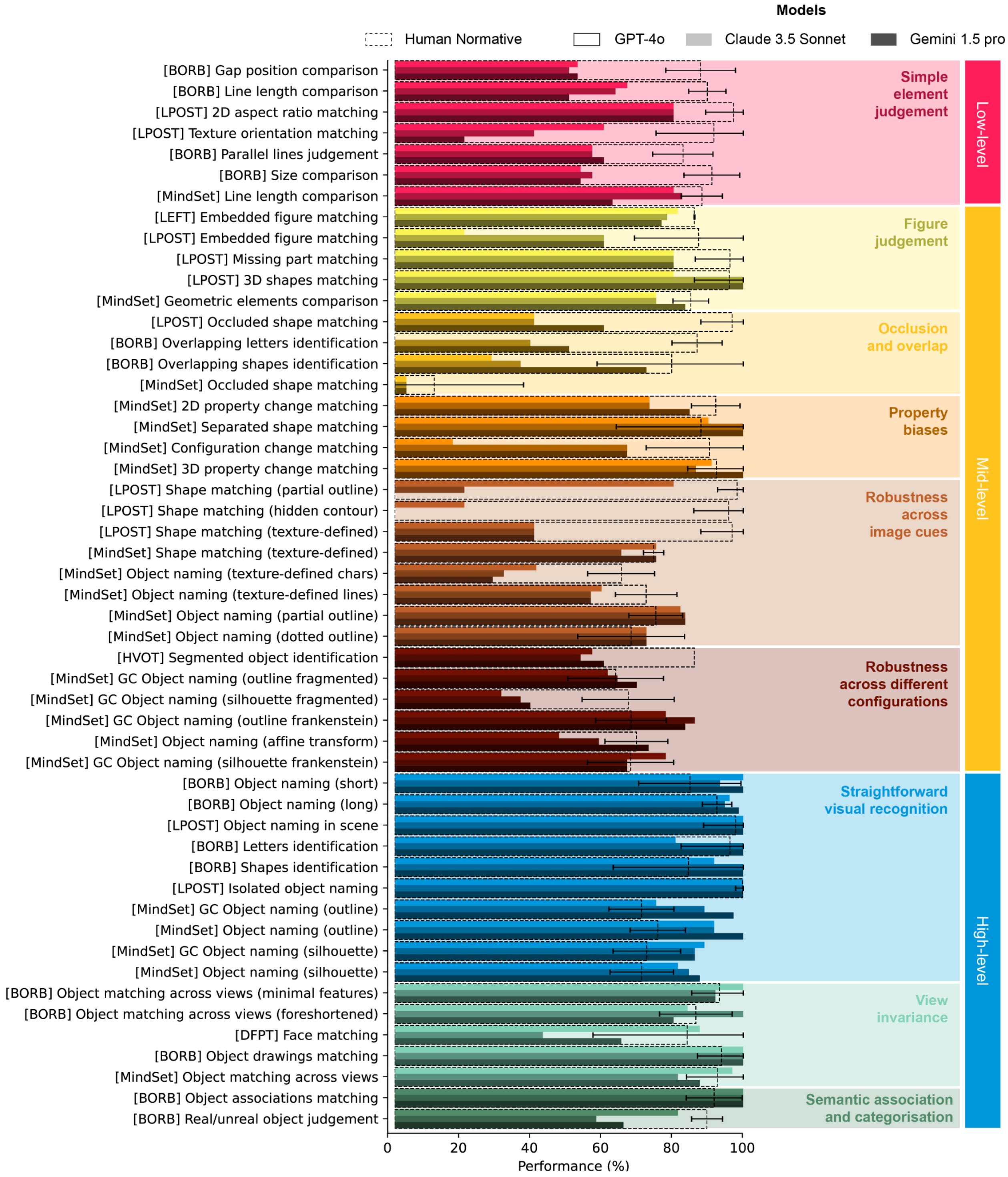

**Supplementary Figure S1. Raw accuracy (% correct) for models and humans across all 51 tasks.** Coloured bars represent the performance of GPT4o's (light), Claude-3.5 Sonnet (dark), Gemini-1.5 pro (darkest), with dashed bars indicating normative human performance. Error bars denote the standard deviation of normative human performance, expressed as a percentage, and are clipped between 0–100%. Tasks are categorized into low-, mid-, or high-level visual assessments and further grouped into nine processes reflecting finger-grained aspects of vision, as in Figure 1 of the main manuscript. A bar without an error bar indicates that the clinical test did not provide a normative standard deviation.